%% file: main.tex
\documentclass[10pt,twocolumn,letterpaper]{article}

\usepackage{cvpr}              
\usepackage{url}
\usepackage{xltabular}
\usepackage{xtab,booktabs}
\usepackage{tabularx}
\usepackage{tabularray}
\usepackage{graphicx}
\usepackage{subcaption}
\usepackage{caption}
\input{preamble}

\definecolor{cvprblue}{rgb}{0.21,0.49,0.74}
\usepackage[pagebackref,breaklinks,colorlinks,citecolor=cvprblue]{hyperref}

\def\paperID{8900} 
\def\confName{CVPR}
\def\confYear{2024}

\title{ChartAssisstant: A Universal Chart Multimodal Language Model via \\ Chart-to-Table Pre-training and Multitask Instruction Tuning}

\author{
Fanqing Meng$^{1, 2}$, Wenqi Shao$^{1\dagger}$, Quanfeng Lu$^{1, 4}$, Peng Gao$^{1}$ \\ Kaipeng Zhang$^{1}$, Yu Qiao$^{1}$, Ping Luo$^{3, 1\dagger}$\\\\
$^{1}$OpenGVLab, Shanghai AI Laboratory \quad $^2$Shanghai Jiao Tong University  \\
$^{3}$The University of Hong Kong \quad $^4$Nanjing University
}

\begin{document}
\maketitle

\renewcommand{\thefootnote}{\fnsymbol{footnote}}
{\let\thefootnote\relax\footnotetext{
\noindent \hspace{-5mm}$\dagger$ Corresponding Authors: shaowenqi@pjlab.org.cn; pluo@cs.hku.edu \\
This work was done when Fanqing Meng and Quanfeng Lu were interning at Shanghai AI Laboratory. 
}   }

\begin{abstract}
Charts play a vital role in data visualization, understanding data patterns, and informed decision-making. However, their unique combination of graphical elements (e.g., bars, lines) and textual components (e.g., labels, legends) poses challenges for general-purpose multimodal models. While vision-language models trained on chart data excel in comprehension, they struggle with generalization. To address these challenges, we propose ChartAssistant, a chart-based vision-language model for universal chart comprehension and reasoning. ChartAssistant leverages ChartSFT, a comprehensive dataset covering diverse chart-related tasks with basic (e.g. bars and pies) and specialized (e.g. radars, and bubbles) chart types. It undergoes a two-stage training process, starting with pre-training on chart-to-table parsing to align chart and text, followed by multitask instruction-following fine-tuning. This approach enables ChartAssistant to achieve competitive performance across various chart tasks. Experimental results demonstrate significant performance gains over the state-of-the-art UniChart and Chartllama method, especially outperforming them on real-world chart data with zero-shot setting. The code and data are available at \href{https://github.com/OpenGVLab/ChartAst}{https://github.com/OpenGVLab/ChartAst}.

\end{abstract}

\section{Introduction}
\label{sec:intro}

\begin{figure}[tb!]
\centering
\includegraphics[width=0.9\linewidth]{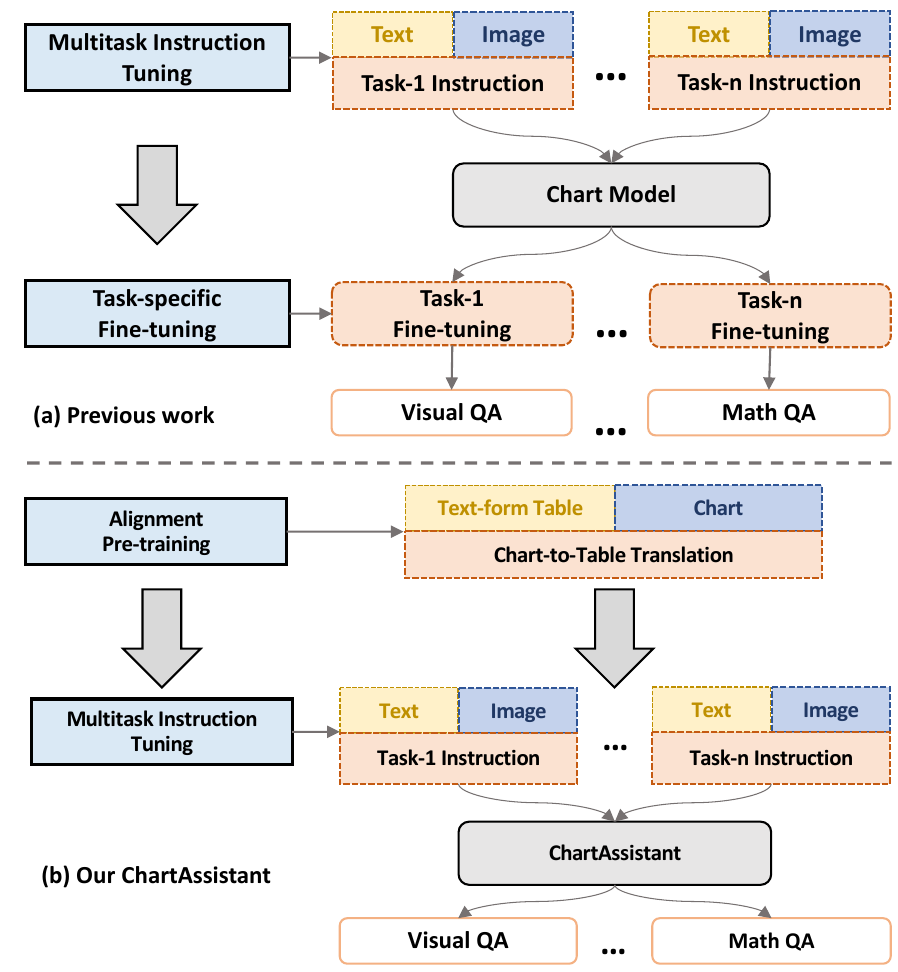} 
\caption{A comparison between previous chart-based models and our proposed ChartAssistant. ChartAssistant first aligns the chart and the text by pre-training on the chart-to-table translation task. After performing multitask instruction tuning, it can solve various downstream tasks.} 
\label{fig:teaser}
\end{figure}

People around the world generate a multitude of charts on a daily basis, including data visualizations for business reports, market analysis, scientific experiments, and data-driven presentations \cite{horn1998visual,hoque2017applying,hoque2022chart}.
Charts are an effective tool for understanding data patterns, such as the distributional properties depicted in histograms and growth trends illustrated in line graphs.
Developing chart learning methods enables the design of machine analysts with enhanced capabilities to solve various chart-related downstream tasks such as chart question answering (QA) \cite{masry2022chartqa,kantharaj2022opencqa,methani2020plotqa}, chart summarization \cite{hsu2021scicap,rahman2022chartsumm}.


However, chart comprehension is challenging due to the intricate visual marks (\emph{e.g.} lines, bars and symbols), implicit numerical information, and complex spatial relationships between elements (\emph{e.g.} axes and labels). Interpreting charts requires specialized knowledge, spatial reasoning, and numerical understanding. The advanced general-purpose multimodal models \cite{zhang2023llama,li2023blip,zhang2023transfer} such as LLaVA \cite{liu2023visual}, trained on natural images, struggle with chart-related tasks due to the specific complexities and relationships unique to charts. Although recent multimodal literate models \cite{lv2023kosmos,lee2023pix2struct} have achieved impressive results in processing various document-level tasks, they still face difficulties in accurately answering chart-related questions.

In pursuit of universal chart reasoning and comprehension, prior works propose pre-training vision-language models on chart-related tasks as shown in Fig.\ref{fig:teaser}(a). For example, 
both MatCha \cite{liu2022matcha} and UniChart \cite{masry2023unichart} undergo multitask instructional tuning and task-specific fine-tuning, exhibiting good performance on several downstream tasks.
%
However, these models still have severe downsides. 
Firstly, they fall short in aligning the chart and the associated structured text-form table, which is essential to interpret the relationships between elements in the chart. Although MatCha \cite{liu2022matcha} underscores the importance of chart-text alignment, it presents poor multitask performance due to limited coverage of chart-related tasks. Secondly, the existing training data \cite{methani2020plotqa, masry2022chartqa} is deficient in image-text annotations aimed at improving the model's comprehension of visual elements and mathematical reasoning, as well as annotated data from the specialized chart types such as box-plots. Due to the above factors, existing chart-based models have poor generalization and require task-specific fine-tuning to achieve promising results on various downstream tasks as illustrated in Fig.\ref{fig:teaser}(a).

To address these challenges, we propose \textbf{ChartAssistant}, a new multimodal model for universal chart comprehension and reasoning. To improve generalization, ChartAssistant is trained on a large-scale chart-specific instruction-tuning benchmark dubbed ChartSFT. The training process involves a two-stage pre-training pipeline which employs chart-to-table pre-training to align the chart and its structured text and then perform joint tuning on multiple chart-related tasks as shown in Fig.\ref{fig:teaser}(b). As a result, our ChartAssistant can achieve good results on various chart-related tasks with a single model.
We implement ChartAssistant with two variants, \emph{i.e.} ChartAst-D and ChartAst-S. ChartAst-D is built upon Donut \cite{kim2021donut}, a lightweight ($260$M parameters) but powerful vision-language model for visual document understanding. 
While ChartAst-S is built upon SPHINX \cite{lin2023sphinx}, a large ($13$B parameters) vision-language model for universal multimodal comprehension. 
Inherited from SPHINX, our ChartAst-S obtains enhanced chart representation by dynamic resolution processing and mixed visual encoders.
Therefore, ChartAst-S offers increased robustness and usability for chart understanding, demonstrating strong performance in various chart-related tasks.

Specifically, we first construct ChartSFT by collecting instruction-following data from various chart-related tasks. To address the limitations of existing chart-based benchmarks \cite{methani2020plotqa, masry2022chartqa,kantharaj2022opencqa}, we introduce several modifications to improve the quality of data annotation: 1) instruction-following data involving various topics for chart-to-table translation is added, which we find helps align the chart and the associated structured text; 2) the chain-of-thought annotations for chart numerical QA task are generated to improve mathematical reasoning abilities \cite{wei2022chain}; 3) the task of chart referring question answering is created to enhance the understanding of visual elements and their relationships \cite{chen2023shikra, yang2023set}; 4) chart with specialized types such as radar and box plot are included to improve the generalization. Overall, ChartSFT encompasses a larger corpus of instruction-following data, incorporates a wider range of chart-related tasks and types, and features more comprehensive data annotations compared to previous benchmarks \cite{masry2022chartqa,methani2020plotqa,kantharaj2022opencqa}.

Before conducting multitask instruction tuning, as done in existing research \cite{masry2023unichart, liu2022matcha}, we start with pre-training ChartAssistant on the chart-to-table translation task as shown in Fig.\ref{fig:teaser}(b). This task involves parsing a chart and generating a Markdown table. It shares similarities with dense captioning for natural images, allowing the model to interpret the elements and relationships within the chart. Similar to the role of image captioning in training multimodal models \cite{liu2023visual, shao2023tiny, xu2023lvlm}, chart-to-table translation facilitates alignment between the chart and its structured text.
Following pre-training, we proceed with multitask instruction tuning using ChartSFT. This two-stage training approach enables ChartAssistant (a single model) to achieve strong performance across a range of chart-related tasks.

The contributions of this paper can be summarized as follows. 
1) We present ChartAssistant, a vision-language model for chart comprehension and reasoning. ChartAssistant is versatile enough to solve various chart-related tasks across a wide range of chart types.
%
2) We build a chart-specific visual instruction-following benchmark dubbed ChartSFT. ChartSFT surpasses existing chart-based benchmarks with its larger instruction-following data corpus, a broader range of tasks and chart types, and more comprehensive data annotations. 
3) Extensive experimental results on various downstream tasks demonstrate that ChartAssistant surpasses the previous SoTA method UniChart \cite{masry2023unichart} by 50.0\%,  28.1\% performance gain on numerical QA and ChartQA, respectively. Notably, ChartAssistant continues to significantly outperform existing Chart-related models in the zero-shot setting, with 29.5\% performance gain on RealCQA \cite{ahmed2023realcqa} compared with Unichart and 23.6\% performance gain on ChartLLM \cite{ko2023natural} compared with Chartllama \cite{han2023chartllama}.

\begin{figure*}[tb!]
\centering
\includegraphics[width=0.9\linewidth]{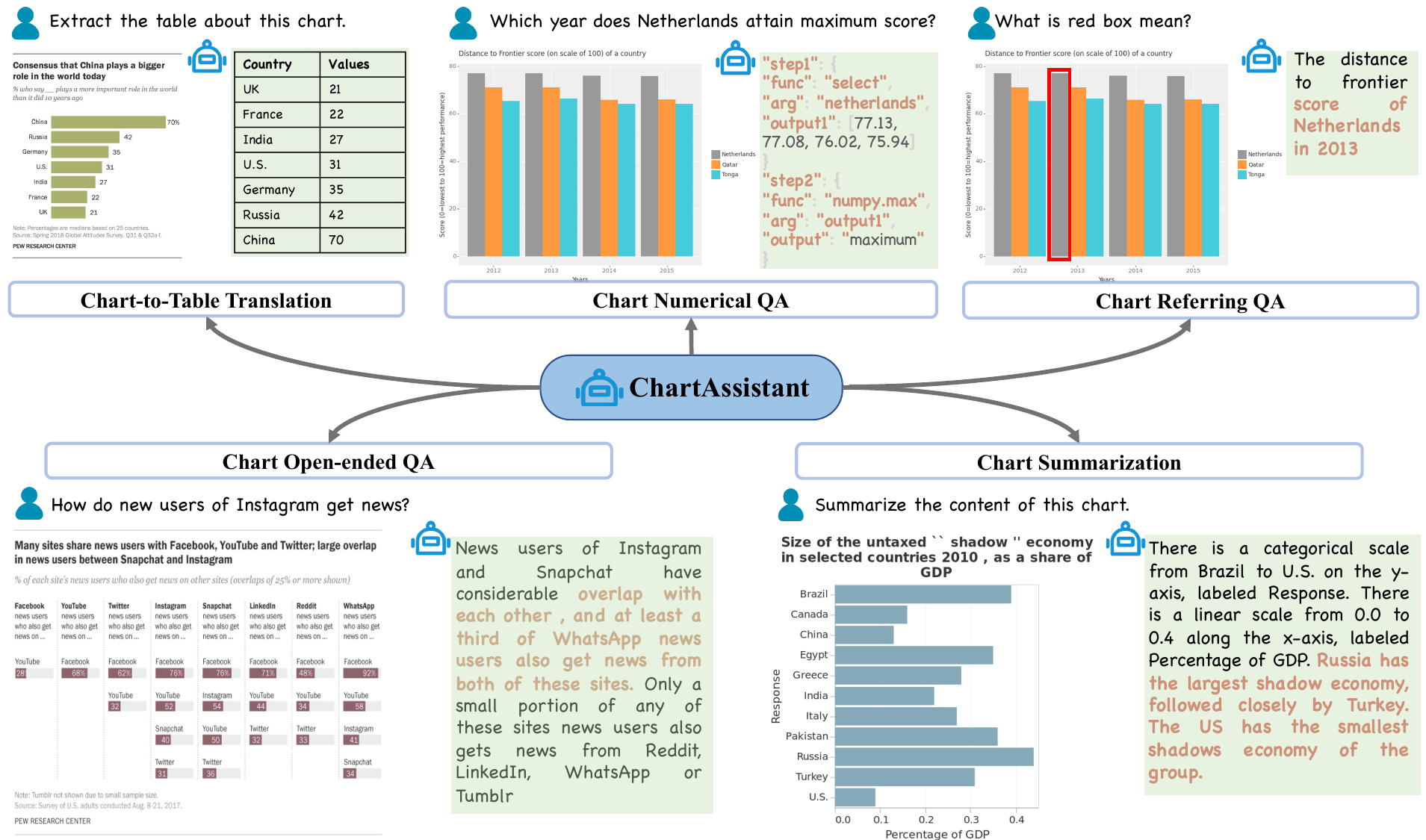} 
\caption{ChartAssistant is pre-trained on vast and various chart-related tasks, and can adeptly perform a range of chart comprehension and reasoning tasks including chart-to-table translation, numerical QA, referring QA, open-ended QA and chart summarization. } 
\label{fig:task}
\end{figure*}

\section{Related Work}

\subsection{Multimodal Foundation Model}
Multimodal foundation models \cite{li2023blip,zhu2023minigpt} mainly focus on natural images, which have shown remarkable progress, advancing in areas like image captioning \cite{vinyals2015show} and visual question answering \cite{vinyals2015show,johnson2017clevr}. SPHINX \cite{lin2023sphinx} leverages LLM and multiple visual encoders to achieve advanced performance on multiple multi-modal tasks. Among these, visual document understanding is a topic of both industrial importance and research challenge. Donut \cite{kim2021donut} proposed an OCR-free Transformer trained in end-to-end manner,which is a powerful document understanding model. Nougat \cite{blecher2023nougat} is fine-tuned on Donut and useful for academic documents understanding.  However, extracting information from real-world images like charts and plots presents unique challenges as compared to natural images or documents. Furthermore, the complexity of queries increases, often involving sophisticated mathematical calculations. As a result, contemporary document models and multimodal foundation models often fall short when tasked with handling chart-related tasks, demonstrating a significant decline in performance \cite{liu2022matcha}.

\subsection{Chart-specific Vision-Language Model}

Some methods modify vision-language models for chart-related tasks \cite{han2023chartllama, liu2023mmc} or develop plugin for LLM to understand the chart \cite{xia2023structchart}. Matcha \cite{liu2022matcha} extends Pix2Struct \cite{lee2023pix2struct} by integrating mathematical reasoning and chart data extraction tasks, excelling at chart question answering and chart summarization. Unichart \cite{masry2023unichart} undergoes multitask instruction tuning for more chart-related tasks, establishing itself as the most versatile and effective chart vision-language model currently available. However, these methods have limitations. Furthermore, these models struggle with mathematical computations, limiting their effectiveness and range of applicable chart types.

Contrastingly, we propose ChartSFT, the most extensive dataset to date, supporting a wide variety of chart tasks and types. We develop ChartAssistant using ChartSFT with a two-stage training strategy, capable of handling diverse chart-related tasks.




\section{ChartSFT}\label{sec:datacoll}

\begin{table*}[t]
\centering
\caption{Summary of utilized datasets and data volumes for each task.}
\resizebox{\textwidth}{!}{
\begin{tabular}{ccccccccccc}
\hline 
ChartQA\cite{masry2022chartqa} & PlotQA\cite{methani2020plotqa} & OpenCQA\cite{kantharaj2022opencqa} & ScigraphQA\cite{li2023scigraphqa} & Vistext\cite{tang2023vistext} & Chart-to-text\cite{kantharaj2022chart} & ChartSumm\cite{rahman2022chartsumm} & arXiv& Data Aug. & SpecializedTypes & Total\\
\hline 
\multicolumn{11}{c}{Chart-to-Table Translation} \\
\hline 
17141 & 224386 & 0 & 0 & 0 & 0 & 0 & 132719 & 220050 & 317662 & 911958 \\
\hline 
\multicolumn{11}{c}{Numerical Question Answering} \\
\hline 
0 & 3997388 & 0 & 0 & 0 & 0 & 0 & 0 & 5318500 & 15178693 & 24494581 \\
\hline 
\multicolumn{11}{c}{Referring Question Answering} \\
\hline 
0 & 0 & 0 & 0 & 0 & 0 & 0 & 0 & 2139567 & 3760275 & 5899842 \\
\hline 
\multicolumn{11}{c}{Open-ended Question Answering} \\
\hline 
30219 & 4362236 & 7724 & 659309 & 0 & 0 & 0 & 408658 & 128105 & 1478952 & 7075203 \\
\hline 
\multicolumn{11}{c}{Chart Summarization} \\
\hline 
0 & 157070 & 7724 & 0 & 12441 & 44096 & 84363 & 0 & 356248 & 419895 & 1006738 \\
\hline
\end{tabular}
}
\label{tab:data}
\end{table*}

\begin{figure}[tb!]
\centering
\includegraphics[width=0.85\linewidth]{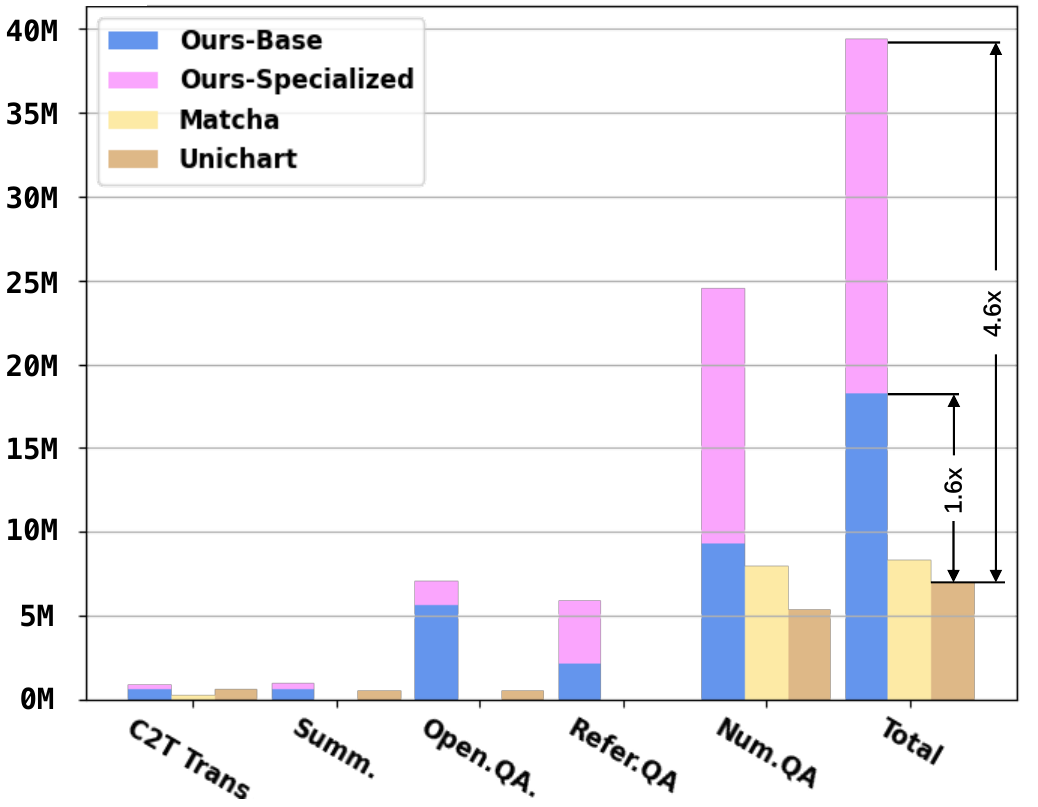} 
\caption{Comparison between ChartSFT and datasets from previous methods. Our dataset surpasses the best previous dataset in UniChart \cite{masry2023unichart} by $4.6$ times in total and supports a greater variety of chart tasks and types. }
\label{fig:data}
\end{figure}

We construct a large-scale chart-specific instruction-tuning benchmark called ChartSFT by collecting data from various tasks. The composition of ChartSFT is shown in Table \ref{tab:data}, as extensively described below. Our ChartSFT consists of $39$M pieces of chart-text annotated data, $4.75$ and $5.62$ times larger than MatCha \cite{liu2022matcha} and UniChart \cite{masry2023unichart}, respectively, as illustrated in Fig.\ref{fig:data}. ChartSFT contains charts with both base and specialized types, as presented in Sec. \ref{sec:base-chart} and Sec. \ref{sec:specilized-chart}, respectively. 

Overall, our ChartSFT encompasses nine types of charts by collecting data from various sources as shown in table \ref{tab:component}. First, most charts with base types including bar, line, dot-line, and pie are collected from several existing datasets \cite{masry2022chartqa, methani2020plotqa, kantharaj2022opencqa, rahman2022chartsumm, li2023scigraphqa, tang2023vistext, kantharaj2022chart}. Second, we also generate some charts with base types from arXiv tables \cite{arXiv} and data augmentation techniques (\emph{e.g.} various APIs and figure parameters). In particular, we use ChatGPT to suggest the proper chart type given each table data from arXiv. Third, we synthesize table data which is appropriate for depicting charts with specialized types.

\begin{table*}[t]
  \centering
  \caption{Chart type distribution of the multitask instruction tuning, we are not including SciGraphQA \cite{li2023scigraphqa} and ChartSumm \cite{rahman2022chartsumm} because these datasets do not contain information about chart types. }
  \label{tab:component}
  \begin{tabular}{@{}lccccccccr@{}}
    \toprule
    Datasets & Bar& Line & Dot-line & Pie & Area & Hist & Radar & Bubble & Box \\
    \midrule
    ChartQA \cite{masry2022chartqa} & 84.8\% & 12.2\% & 0.0\% & 3.0\% & 0.0\% & 0.0\% & 0.0\% & 0.0\% & 0.0\% \\
    PlotQA \cite{methani2020plotqa} & 67.0\% & 16.5\% & 16.5\% & 0.0\% & 0.0\% & 0.0\% & 0.0\% & 0.0\% & 0.0\% \\
    OpenCQA \cite{kantharaj2022opencqa} & 71.7\% & 24.6\% & 0.6\% & 3.0\% & 0.1\% & 0.0\% & 0.0\% & 0.0\% & 0.0\% \\
    Vistext \cite{tang2023vistext} & 50.1\% & 24.2\% & 0.0\% & 0.0\% & 25.6\% & 0.0\% & 0.0\% & 0.0\% & 0.0\% \\
    Chart-to-text \cite{kantharaj2022chart} & 82.8\% & 13.6\% & 1.6\% & 0.0\% & 0.0\% & 0.0\% & 0.0\% & 0.0\% & 0.0\% \\
    arXiv& 71.6\% & 17.1\% & 11.3\% & 0.0\% & 0.0\% & 0.0\% & 0.0\% & 0.0\% & 0.0\%  \\
    Data Aug. & 56.5\% & 17.0\% & 11.5\% & 15.0\% & 0.0\% & 0.0\% & 0.0\% & 0.0\% & 0.0\% \\
    Synthetic & 0.0\% & 0.0\% & 0.0\% & 0.0\% & 23.1\% & 25.8\% & 20.9\% & 19.1\% & 11.1\% \\
    \midrule
    Total & 44.3\% & 11.3\% & 8.0\% & 3.6\% & 7.8\% & 8.4\% & 6.8\% & 6.2\% & 3.6\% \\
    \bottomrule
  \end{tabular}
\end{table*}


\subsection{Chart with Base Types}\label{sec:base-chart}
We collect instruction-following data with base chart types (\emph{i.e.} bars, lines, dot-lines, and pies) from $5$ chart-rated tasks, including chart-to-table translation, chart numerical QA, chart referring QA, chart open-ended QA, and chart summarization as shown in Fig.\ref{fig:task}. Instead of directly utilizing existing chart-based benchmarks, we introduce several modifications to improve the data annotation quality.
For each task, we present the details of data collection as follows.

\subsubsection{Chart-to-Table Translation}\label{sec:chart2table-trans}
The task of chart-to-table translation aims at parsing a chart into its underlying data table in text form. Pre-training with chart-to-table translation enables our ChartAssistant to comprehend the chart's elements and their relationships, facilitating alignment of the chart and its underlying structured text. 

\textbf{Data Collection.} We collect $17141$ and $224386$ pieces of chart-text data from ChartQA and PlotQA for chart-to-table translation. However, these benchmarks vary little in chart styles and involve limited topics. We propose two strategies to address the issue. 
\begin{itemize}
    \item \textbf{More Chart Styles.} We re-plot the chart with diverse visualization tools for tables in ChartQA and PlotQA. Specifically, we utilize $5$ APIs in Python, including ggplot, plotly, matplotlib, seaborn, and pyecharts, along with over $20$ variations in parameters color, size, font type, background, and more. After style augmentation,  $220050$ pieces of chart-text data are created for chart-to-table translation from PlotQA, respectively. 
    \item  \textbf{Table from arXiv Papers.} We collect more real table data to increase the topic diversity.  To this end, we crawl $1301932$ papers involving various topics such as computer science, biology, finance, and more from arXiv platform \cite{arXiv}. For each paper, we extract the table from the source LaTeX code where table data can be localized in the table environment. We employ ChatGPT \cite{ouyang2022training} to transform the latex table into the markdown table. We also make the chart in a specific base type (\emph{e.g.} pies) by following ChatGPT's suggestion. We find that ChatGPT works well to generate text in the target format and give appropriate advice for chart types. There are $132719$ pieces of chart-text data obtained from the arXiv.
\end{itemize}

\subsubsection{Chart Numerical Question Answering}\label{sec:numerical-QA}

Chart numerical QA targets at responding to the request about mathematical reasoning given a chart.  It requires an accurate understanding of the chart, as well as reasoning and math calculation abilities.

\textbf{Data Collection.} The data for numerical QA mainly comes from the PlotQA benchmark. However, PlotQA generates numerical QA data from $40$ templates with limited types of questions and direct final answers, resulting in poor generalization and math reasoning. with our proposed two strategies to improve the data quality below,  more than 24M QA pairs are collected.

\begin{table}[]
    \centering
        \caption{Comparison of templates for numerical QA between PlotQA and our ChartSFT. `Num.' denotes the number of templates. We use $4$ statistics to measure the complexity of templates, including `Len.', `COT Steps' and `Fun.'. They denote the average token length, the number of steps in COT annotation, and how many kinds of functions are needed to obtain the final answer, respectively. Besides templates in PlotQA, ChartSFT newly created $61$ templates for numerical QA with higher complexity.}
         \label{tab:comparison-template}
    \begin{tabular}{c|c c c c }
    \toprule
                   & Num. & Len. & COT Steps &Func.   \\
                   \hline
        PlotQA     &  40   &32.83 &3.48  & 2.95    \\
        ChartSFT &  61 (101)   &39.54 &5.02  & 3.90 \\
        \bottomrule
    \end{tabular}
\end{table}

\begin{itemize}
    \item \textbf{More Templates.} We create $101$ templates to generate numerical QA questions automatically involving various types of questions with complex calculations. Here is one template for analyzing the correlation between two items: `Across all $<$plural form of X label$>$, are the $<$Y label$>$ values of $<$legend label1$>$ and $<$legend label2$>$ negatively correlated?' The comparison between templates in our ChartAssistant and PlotQA is provided in Table \ref{tab:comparison-template} where we can see that our improved templates encompass larger token lengths and more complex calculations. We present all templates in the Appendix Sec.A.
    \item  \textbf{Chain-of-Though (COT) Annotations.} Instead of utilizing the final answer as the response annotation, we generate COT annotation for the final answer, which has been proven to improve the model's mathematical reasoning ability \cite{wei2022chain}. We first define a set of available functions to segment the problem's solution into smaller steps, each encompassing function calls and parameters. These steps are then organized into a JSON-formatted text. As shown in Fig.\ref{fig:task}, the maximum extraction problem is decomposed into a step of data retrieval and a step of maximum calculation. When computing the answers, the backend executes the calculations by following the ordered function calls within the text. This approach not only enhances reasoning ability but also mitigates calculation errors. 
\end{itemize}

\subsubsection{Chart Referring Question Answering}\label{sec:referring-QA}
We create a new task for chart named referring question answering, considering that users may utilize a set of marks to denote some pieces to their interest in the chart as shown in Fig.\ref{fig:task}. Note that referring question answering with a bounding box has been explored in general-purpose multimodal models such as GPT4ROI \cite{zhang2023gpt4roi} and Shikra \cite{chen2023shikra} where the referential QA has been shown to benefit comprehending spatial relationships. The task of referring QA is expected to enhance the understanding of visual elements and their relationship in the chart.

\textbf{Data Collection.} We extend a part of COT annotations for numerical QA in Sec.\ref{sec:numerical-QA} to the task of Referring QA. Three steps are conducted to produce referring QA pairs with diverse patterns. i) The color, size, and width are randomly selected to make the mark. ii) We use several marks such as an arrow and a bounding box to refer to an item in the chart. iii) Multiple marks can be depicted in the same chart to describe the relationships between elements. Overall, we collect $5899842$ pieces of data for the chart referring QA.

\subsubsection{Chart Open-ended QA}
Chart open-ended QA (OpenQA) deals with open-ended questions regarding charts as illustrated in Fig.\ref{fig:task}. It requires both low-level Chart comprehension and high-level reasoning abilities.

\textbf{Data Collection.}  We collect data from existing benchmarks, such as plotQA \cite{methani2020plotqa}, ChartQA \cite{masry2022chartqa}, OpenCQA \cite{kantharaj2022opencqa} and ScigraphQA \cite{li2023scigraphqa}. We further introduce our collected table data from arXiv  in Sec.\ref{sec:chart2table-trans} for this task. 
    
\textbf{Open-ended QA data by ChatGPT.} Other than tabular data crawled in Sec.\ref{sec:chart2table-trans}, we extract corresponding captions, and the first paragraph describing the table from the source code of the paper. By utilizing ChatGPT, we generate $3$ open-ended QA pairs for each table by feeding the table and the descriptive information.


By putting the above benchmarks together, our ChartSFT covers diverse topics for Open-ended QA. In total, there are $7075243$ pieces of data for this task.

\begin{figure*}[tb!]
\centering
\includegraphics[width=0.95\linewidth]{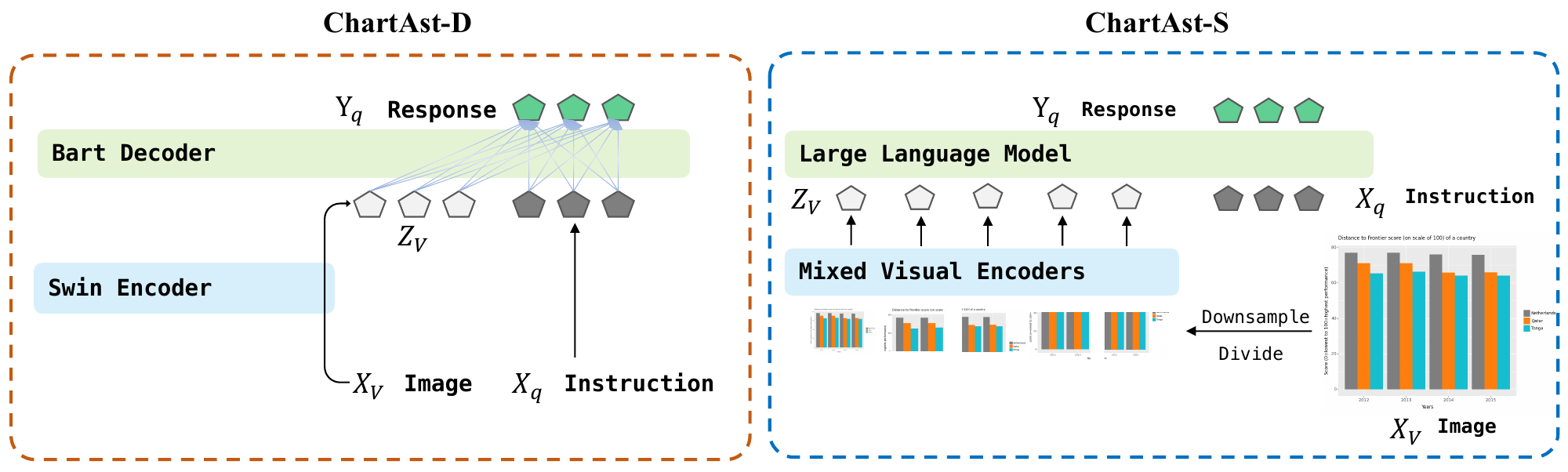} 
\caption{ChartAst-D and ChartAst-S network architecture. }
\label{fig:model}
\end{figure*}

\subsubsection{Chart Summarization}
Chart Summarization is a vital task aimed at generating concise and informative summaries for various types of charts, which has been studied extensively \cite{herdade2019image,tang2023vistext,kantharaj2022chart}.

\textbf{Data Collection.} We collected a substantial amount of existing open-source datasets \cite{tang2023vistext,kantharaj2022chart,rahman2022chartsumm,kantharaj2022opencqa}, but the scale is still not sufficient. Therefore, we further incorporate a large-scale chart summarization dataset generated through Knowledge Distillation by Unichart \cite{masry2023unichart} into our training process. There are $1006738$ pieces of data for the chart summarization task.

\subsection{Charts with Specialized Types}\label{sec:specilized-chart}
Previous chart-based models have exhibited poor performance when dealing with specialized chart types, such as radar, area, histogram, bubble, and box-plot. To enhance the model's generalization capabilities, we have trained our ChartAssistant on these charts with specialized types. To overcome the challenge of obtaining large-scale real-world chart data, we have employed synthetic data generation techniques. For more detailed information, please refer to Appendix Sec. \ref{sec:appen-chartsft-spe}. Through this approach, we can obtain a substantial and diverse collection of complex charts across these specialized types.

\section{Our ChartAssistant}

\subsection{Architecture}

The key to completing the tasks related to charts lies in accurately understanding the content of the charts. As shown in Fig. \ref{fig:model}, we implement ChartAssistant with two variants, \emph{i.e.} ChartAst-D and ChartAst-S, which have 260M and 13B parameters in total. In addition, their input image resolutions are $224\times 224$ and $448\times 448$, respectively. Both ChartAst-D and ChartAst-S perform well in many chart-related tasks. But ChartAst-D has a smaller size and ChartAst-S enjoys better generalization.


\textbf{ChartAst-D} is a vision-language model for chart understanding built upon Donut \cite{kim2022ocr}. It consists of a visual encoder Swin-Base \cite{liu2021swin} and a textual BART decoder \cite{lewis2019bart}. For an input image $X_{V}$, the visual encoder employs fixed-sized non-overlapping windows to divide the image and performs self-attention layers to consolidate information across these windows, which transforms the image into a set of tokens $Z_V=\left\{\mathbf{z}_{i} \mid \mathbf{z}_{i} \in \mathbb{R}^{d}, 1 \leq i \leq n\right\}$, where $n$ is encoded token length and $d$ is the token size. 
By taking $Z_V$ as key and value and tokens of text instruction $X_q$ as the query, the BART decoder generates the corresponding response $Y_q = \left(\mathbf{y}_{i}\right)_{i=1}^{m}$, and $m$ is the length of responses.

\textbf{ChartAst-S} is a large vision-language model for chart understanding built upon Sphinx \cite{lin2023sphinx}. For high-resolution images, it preserves the original information through sampling and partitioning methods, ensuring greater fidelity to the image content. Moreover, Sphinx leverages the abundant prior knowledge of LLM \cite{touvron2023llama} to handle various tasks such as visual question answering and image summarization. Specifically, for an input image $X_{V}$. ChartAst-S incorporates multiple visual encoders to extract more informative visual features $Z_V$, such as DINOv2 \cite{oquab2023dinov2}, CLIP \cite{radford2021learning}, and ConvNeXt \cite{woo2023convnext}. Unlike ChartAst-D where visual tokens are involved in a language decoder with a cross-attention module, ChartAst-S directly appends visual tokens to the text tokens $X_q$. The merged tokens are then fed into the LLM to generate the response. Thanks to the intricate design of the visual encoder and the powerful reasoning ability of LLM, ChartAst-D generalizes well in various real-world chart-related applications.

\subsection{Training}

In our ChartSFT, we have a corresponding instruction $X_{q}$ and response $Y_{q}$ for each image $X_{V}$. We input these image-text pairs into the model. The objective is to minimize the cross-entropy loss of predicting the next token.
To improve the generalization in various downstream tasks, we adopt a two-stage training pipeline to train our ChartAst-D and ChartAst-S below. Charts are special images that visualize the data and underlying relationships between elements in the chart. Understanding the numerical values and their meanings is a prerequisite for completing downstream tasks related to charts. Therefore, we employ Chart-to-Table translation as a pre-training task, aiming to enable the model to understand the correspondence between charts and tables, which has also been utilized as a part of a pre-training task in MMC \cite{liu2023mmc} and Matcha \cite{liu2022matcha}.

\textbf{Stage \uppercase\expandafter{\romannumeral1}: Pretraining on Chart-to-table Translation.}  Given a chart $X_V^{c2t}$, our goal is to convert the chart into a text-form table $Y_{q}^{c2t}$ under the instruction $X_q^{c2t}$. Here the superscript $c2t$ indicates the instruction-following data comes from the task of chart-to-table translation. Our training loss function for Stage \uppercase\expandafter{\romannumeral1} is given by
\begin{equation}\label{eq:c2t-pretraining}
    \mathcal{L}^{\mathrm{Stage1}} = -\sum_{i=1}^m \log P_\theta(Y_{q,i}^{c2t}|X_V^{c2t}, X_q^{c2t},Y_{q,<i}^{c2t}),
\end{equation}
where $Y_{q,<i}^{c2t}$ are all the response tokens before the current prediction token $Y_{q,i}^{c2t}$. $\theta$ are the learnable weights initialized from the pre-trained weights of the Donut model \cite{kim2021donut}.

By the pre-training in Eqn. (\ref{eq:c2t-pretraining}), we align the chart with its structured text-form table, enabling the model to comprehend elements in charts and their relationships. We show that this strategy better serves the multitask instruction tuning in Sec.\ref{sec:AblationStudy}.

\textbf{Stage \uppercase\expandafter{\romannumeral2}: Multitask Instruction Tuning.}  
In this stage, we put all the instruction-following data together from five tasks in our ChartSFT. We employ a single model to solve all the tasks.  Our training loss function for Stage \uppercase\expandafter{\romannumeral2} is given by
\begin{equation}\label{eq:c2t-multitask}
    \mathcal{L}^{\mathrm{Stage2}} = -\sum_{k\in\Omega}\sum_{i=1}^m \log P_\theta(Y_{q,i}^{k}|X_V^{k}, X_q^{k},Y_{q,<i}^{k}),
\end{equation}
where $\Omega$ is the set of instruction-following data from all tasks in ChartSFT and $\theta$ are the learnable weights initialized from the checkpoint in the Stage \uppercase\expandafter{\romannumeral1}. During training, we sample the data from each task with certain proportions as provided in our experimental setup in Appendix Sec.B. By multitask instructional tuning, our ChartAssistant exhibits strong performance on all the tasks.

\section{Experiment}
we present our experimental setup in Appendix Sec.B, where we indicate the training details. After that, we provide an overview of the selected baselines and evaluation details in Sec.\ref{Baselines} and demonstrate the superior effectiveness of our method through extensive experiments in Sec.\ref{Results} .


\subsection{Baselines and Evaluation}\label{Baselines}

\textbf{Evaluation.}  We assess the performance of ChartAssistant across various tasks and datasets. Following the evaluation of Unichart \cite{masry2023unichart}, we utilize Chart-to-text \cite{kantharaj2022chart} for evaluating chart summarization task,  and OpenCQA \cite{kantharaj2022opencqa} and ChartQA \cite{masry2022chartqa} for open-ended question answering task. To evaluate numerical question answering and referring question answering, we sample test sets from the datasets constructed in Sec.\ref{sec:numerical-QA} and  Sec.\ref{sec:referring-QA} called MathQA and ReferQA. Lastly, we conduct separate evaluations on base type and specialized type charts to highlight the superior performance of our method more explicitly. We put a detailed description of the dataset in Appendix Sec.B.
%

\textbf{Metrics.} For evaluating ChartQA, MathQA, and ReferQA, we adopt the approach used in previous studies \cite{liu2022matcha,masry2023unichart}, which considers relaxed correctness (allowing for an exact match with tolerance for a 5\% numerical error). As for Chart-to-Text and OpenCQA, we employ BLEU as the evaluation metric following previous works \cite{liu2022matcha,masry2023unichart}. For chart-to-table translation, we use $RMS_{F1}$ from DePlot \cite{liu2022deplot}.

\textbf{Baselines.} We choose SPHINX \cite{lin2023sphinx}, Blip2-flant5-xl \cite{li2023blip}, Qwen-VL \cite{bai2023qwen}, ChartLLaMa \cite{han2023chartllama}, Unichart \cite{masry2023unichart}, Matcha \cite{liu2022matcha}, Pix2Struct \cite{lee2023pix2struct}, T5 \cite{raffel2020exploring} and Chart-T5 \cite{zhou2023enhanced} as baselines. ChartLLama and Unichart are the current state-of-the-art models that handles the maximum number of chart tasks and delivers the best overall performance. Besides, Unichart also considers the open-ended QA task. Matcha outperforms previous models in mathematical calculations. Pix2Struct and Donut stands out as an excellent document understanding model. We fine-tune these document models on the train set of the respective evaluation datasets and present the results. T5 is a text-to-text model and needs OCR-based system to extract the data table from the chart image, Chart-T5 is a model modified from T5 for chart-related tasks. We use the results from Unichart \cite{masry2023unichart} for them. SPHINX \cite{lin2023sphinx}, Blip2\cite{li2023blip} and Qwen-VL \cite{bai2023qwen} are all commonly used large vision-language models at present. We observe that these models underperform in processing Chart tasks. Finally, Chartllama, utilizing LLaVA for training on Chart data, demonstrates superior performance in Chart tasks. Therefore, we only compare with Chartllama.


\begin{table}[!t]
 \centering
  \caption{\small {A comparison of the results of ChartAssistant with the existing Chart model on five tasks with base type charts, which shows that ChartAssistant is ahead of the rest of the models on all tasks. Bold indicates best results, italics indicate that the model is not trained on this task.
 }
 \vspace{-3mm}
 }
 
 \scalebox{0.53}{{\begin{tabular}{ll|cccccccc}
  
  \toprule
  
   & & \multicolumn{2}{c}{ChartQA} & 
  \multicolumn{2}{c}{Chart-to-Text} & \multicolumn{1}{c}{Chart-to-Table}   \\ 
  \cmidrule(lr){3-4} \cmidrule(lr){5-6} \cmidrule(lr){7-7}
  
   Model & Size & aug. & human & Pew & Statista & ChartQA & OpenCQA &MathQA  & ReferQA \\

  \midrule

  T5 & 223M & 41.0 & 25.1 
  & 10.5 & 35.3 & - & 9.3 & - & -  \\ 

  Chart-T5 & 400M & 74.4 & 31.8 
  & 9.10 & 37.5 & - & - & - & - \\

   Donut & 260M  & 78.1 & 29.8 & 
  7.2 & 38.2 & 87.4 & 13.1 & 36.3 & \textit{6.2} \\ 

  Pix2Struct& 300M  & 81.6 & 30.5 & 
  10.3 & 38.0 & 85.9 & 12.7 & 35.6 & \textit{5.8} \\

    SPHINX& 13B  & \textit{11.3} & \textit{21.7} & 
  \textit{3.2} & \textit{4.1} & \textit{9.4} & \textit{5.9} & \textit{4.4} & \textit{7.2} \\ 
    Qwen & 9.6B  & 78.9 & 44.3 
  & \textit{0.5} & \textit{2.6} & - & \textit{1.3} & \textit{4.8} & \textit{4.9} \\

  Blip2 & 4B  & \textit{1.4} & \textit{7.8}
  & \textit{0.2} & \textit{0.8} & - & \textit{1.7} & \textit{6.4} & \textit{0.4} \\ 
  
  Matcha& 300M  & 88.9 & 38.8 
  & 12.2 & 39.4 & 89.6 & \textit{6.5} & 57.8 & \textit{8.3} \\

  Unichart& 260M & 87.8 & 43.9 
  & 12.5 & 38.1 & 91.1 & 14.8 & 23.9 & \textit{11.9} \\ 
    
  ChartLLaMa & 13B  & 90.4 & 48.9 
  & 14.2 & 40.7 & 90.0 & \textit{4.7} & \textit{5.8} & \textit{9.9} \\ \midrule
  
  \textbf{ChartAst-D}& 260M & 91.3 & 45.3
  & 14.0 & 40.2 & \textbf{92.0} & 14.9 & 72.1 & 64.2 \\ 

  \textbf{ChartAst-S} &13B& \textbf{93.9} & \textbf{65.9} 
  & \textbf{15.2} & \textbf{41.0} & 91.6 & \textbf{15.5} & \textbf{73.9} & \textbf{67.9} \\
  \bottomrule

 \end{tabular}}
 }
 \vspace{-3mm}

 \label{tab:performance-base}
\end{table}

\subsection{Main Results}\label{Results}

\textbf{Base type charts.} In table \ref{tab:performance-base}, we present a comprehensive summary of ChartAssistant's performance with base type charts across chart-related tasks. It demonstrates that ChartAssistant consistently outperforms the baseline across all tasks. In partivular, we surpass the current leading methods by 17\% and 2.5\% on ChartQA-human and ChartQA-augment, respectively. Besides, we find that most existing models struggle with numerical question answering, while the adaptation of COT answer significantly enhances performance, demonstrating a substantial 16.1\% improvement with Matcha. Notably, existing models are currently almost unable to effectively handle the chart referring question answering task. At last, In summation, our model is the top performer across all chart-related tasks.  It is important to note that both Unichart and Matcha's results are given after fine-tuning on the training set of the test dataset, whereas ChartAssistant's results are obtained using a single model after training is complete, except for Chart-to-Text, because this dataset contain too few groundtruth references, which means that the results must be very close to the reference targets to achieve high scores when evaluating using BLEU-4 \cite{han2023chartllama}.

\textbf{Specialized type charts.}  Following the similar training strategy shown in Fig.\ref{fig:task}, we fine-tune ChartAssistant on chart data of specialized types. As depicted in table \ref{tab:res_othertype}, compared to the current chart-specific vision-language models, none of them can generalize effectively to specialized types of charts due to lack of these training data. ChartAssistant demonstrates an absolute advantage in all five tasks related to specialized types of charts compared to them.

\begin{table}[!t]
 \centering
\caption{A comparison of the results of ChartAssistant with other chart-specific models on five tasks with specialized type charts. Use BLEU to evaluate summarization and open-ended QAs.}
 
 \scalebox{0.78}{\begin{tabular}{l|cccccc}
  
  \toprule

   Model & C2T Trans. & Summ.& Open.QA & Num.QA & Refer.QA \\

  \midrule

  Matcha & 17.1 & 6.3
  & 5.1 & 7.2 & - \\ 
                          
  Unichart & 18.4 & 6.3
  & 5.4 & 5.9 & -  \\ 

  Chartllama & 19.4 &  9.2
  & 8.4 & 2.4 & -  \\ \midrule

  \textbf{ChartAst-D} & 68.3 & 19.7
  & 25.7 & 42.5 & 65.2\\
  \textbf{ChartAst-S} & \textbf{75.6} & \textbf{22.0} 
  & \textbf{27.8} & \textbf{49.8} & \textbf{68.4} \\ \bottomrule
  \end{tabular}
    }

    \label{tab:res_othertype}
\end{table}

\section{Zero-shot Study} \label{sec:zero-shot}
In addition to outperforming the current best methods on common datasets such as ChartQA and Chart-to-text, ChartAssistant-S demonstrates its excellence. To validate the model's generalizability, it is necessary to test on samples not included in the training set. For this purpose, we have collected data from StructChart \cite{xia2023structchart}, RealCQA \cite{ahmed2023realcqa}, and ChartLLM \cite{ko2023natural} 
for tasks like chart-to-table translation, chart-based question answering, and summarization. The results indicate that ChartAssistant exhibits superior zero-shot performance across all tasks, surpassing current methods. We also present a selection of comparative examples in the supplementary materials for visualization.
For evaluation, RealCQA uses accuracy within a 5\% error margin, ChartLLM employs GPT-4 scoring used in Chartllama \cite{han2023chartllama}, while StructChart is evaluated using $RMS_{F_1}$ metrics.  As shown in table \ref{tab:zero-shot}, we find that in the zero-shot setting, Chartllama performs poorly in precise numerical question answering but excels in summarization tasks. We attribute this to the robust language capabilities of LLM. On the other hand, ChartAssistant surpasses existing models in tasks such as precise numerical question answering in OCR and summarization, which involves generating long texts. Furthermore, we observe that if the model's decoder is not powerful enough, errors are more likely to occur in the zero-shot setting when tasked with generating long text outputs, such as in summarization or providing answers in COT format. The use of Large Language Models (LLMs) can significantly alleviate this issue.

\begin{table}[!t]
 \centering
     \caption{In comparison with other chart-related multimodal models in a zero-shot setting, ChartAssistant-S significantly outperforms existing models across all tasks in the zero-shot scenario. }
 \scalebox{0.75}{
 \begin{tabular}{l|cccc}
  
  \toprule
  
   & \multicolumn{2}{c}{RealQA}   \\ 
  \cmidrule(lr){2-3} \cmidrule(lr){4-4} \cmidrule(lr){5-5}
  
   Model & Math & Extract & ChartLLM & StructChart  \\

  \midrule

  Unichart & 13.0 & 33.0
  & 11 & 41.5   \\

  Matcha & 16.0 & 27.5 
  & 11 & 23.3  \\ 

Chartllama & 10.0 & 13.0 
  & 55 & 38.3  \\ \midrule
  \textbf{ChartAst-D} & 15.0 & 36.0
  & 13 & 39.4 \\
  \textbf{ChartAst-S} & \textbf{32.0} & \textbf{43.5}
  & \textbf{68} & \textbf{45.3}   \\ \bottomrule
  \end{tabular}
  }
    \label{tab:zero-shot}
\end{table}

\section{Ablation Study} \label{sec:AblationStudy}
We thoroughly analyze the key aspects of our approach. We first consider the significance of alignment pre-training and the referring question answering task. Furthermore, we evaluate the impact of the COT answer and each task on the effectiveness of our approach. We put more experiments in Appendix Sec.C, including the significance of arXiv data and generation of equivalent questions on the effectiveness and robustness of our approach. We adopt ChartAst-D to illustrate the superiority of our designed ChartSFT, as well as to emphasize the importance of the training strategy.

\begin{table}[!t]
 \centering
     \caption{A comparison of the results of ChartAssistant with its variants on five tasks with base type charts, which indicates that the alignment pretraining and the referring question answering task play a crucial role in enhancing the overall performance. }
 \scalebox{0.55}{\begin{tabular}{l|cccccccc}
  
  \toprule
  
   & \multicolumn{2}{c}{ChartQA} & 
  \multicolumn{2}{c}{Chart-to-Text} & \multicolumn{1}{c}{Chart-to-Table}   \\ 
  \cmidrule(lr){2-3} \cmidrule(lr){4-5} \cmidrule(lr){6-6}
  
   Model & aug. & human & Pew & Statista & ChartQA & OpenCQA &MathQA  & ReferQA \\

  \midrule

Ours-D w/o align & 89.0 & 42.1 
  & 13.7 & 38.3 & 89.5 & 14.3 & 62.3 & 60.1  \\

  Ours-D w/o refer & 89.2 & 41.2
  & 14.0 & 38.6 & 90.7 & 14.6 & 60.2 & - \\ \midrule

  \textbf{Ours-D} & \textbf{91.3} & \textbf{45.3} 
  & \textbf{14.0} & \textbf{40.2} &\textbf{92.0} & \textbf{14.9} & \textbf{72.1} & \textbf{64.2} \\ \bottomrule
  \end{tabular}
    }

    \label{tab:ablationocr}
\end{table}


\textbf{The impact of alignment pretraining.} We initially validate the importance of alignment pretraining. We ensure that the "Ours w/o align" version of the model is trained for the same number of iterations as the full ChartAssistant model. Table \ref{tab:ablationocr} shows that using only multitask instruction tuning falls considerably behind two-stage training strategies. Exact numerical recognition greatly influences mathematical calculation accuracy, leading to a 9.8\% and 3.2\% performance drop for MathQA and ChartQA-human tasks. We think alignment pre-training, which allows the model to learn chart-table correlations, helps the model better adapt during multitask instruction tuning than handling these processes separately \cite{liu2023visual}.

\textbf{The impact of referring question answering task.} In our experiments, we have observed that integrating referring question answering into multitask instruction tuning training can enhance the model's performance in other tasks. As shown in table \ref{tab:ablationocr}, incorporating the referring question answering task leads to improvements across almost all tasks, particularly in tasks requiring mathematical reasoning. For instance, the average performance in ChartQA improves by 3.1\%, and in MathQA, it improves by 11.9\%. We believe that this task strengthens the model’s ability to understand the visual elements and their relationship in the chart, which contributing to overall performance enhancement \cite{zhang2023gpt4roi,chen2023shikra}.

\textbf{The impact of arXiv data.} we conduct experiments by excluding the arXiv data at two distinct stages: the alignment pre-training (stage 1), and the multitask instruction tuning (stage 2). As shown in table \ref{tab:ablationarxiv}, it demonstrates that the arXiv dataset significantly assists the model in aligning charts with tables, thereby improving the performance across various tasks. We believe this is due to the fact that in comparison to existing chart-to-table translation datasets, the arXiv dataset boasts more diversity in terms of style and context; Besides, the open-ended question-answering task contributed by the arXiv dataset is proved to be pivotal for the multitask instruction tuning. We note that the removal of this leads to a drop in the performance of all tasks, most notably math QA and the referring QA. The possible reason for this is because the context and diverse meanings of the arXiv dataset contribute to higher quality question and answering pairs. Therefore, it better promotes multitask tuning.



\begin{table}[!t]
 \centering
     \caption{A comparison of the results of ChartAssistant without arXiv dataset on five tasks with base type charts, which indicates that the arXiv dataset significantly improve the performance of the alignment pre-training and mulittask instruction tuning. }
 \scalebox{0.55}{\begin{tabular}{l|cccccccc}
  
  \toprule
  
   & \multicolumn{2}{c}{ChartQA} & 
  \multicolumn{2}{c}{Chart-to-Text} & \multicolumn{1}{c}{Chart-to-Table}   \\ 
  \cmidrule(lr){2-3} \cmidrule(lr){4-5} \cmidrule(lr){6-6}
  
   Model & aug. & human & Pew & Statista & ChartQA & OpenCQA & MathQA  & ReferQA \\

  \midrule

  stage1 w/o arXiv & 89.9 & 43.7 
  & 13.8 & 39.1 & 91.1 & 14.5 & 64.1 & 61.1  \\

  stage2 w/o arXiv & 89.7 & 42.6
  & 12.6 & 37.5 & 91.3 & 13.2 & 56.7 & 56.4 \\ \midrule

  \textbf{Ours-D} & \textbf{91.3} & \textbf{45.3} 
  & \textbf{14.0} & \textbf{40.2} &\textbf{92.0} & \textbf{14.9} & \textbf{72.1} & \textbf{64.2} \\ \bottomrule
  \end{tabular}
    }

    \label{tab:ablationarxiv}
\end{table}

\begin{figure}[tb!]
\centering
\includegraphics[width=0.8\linewidth]{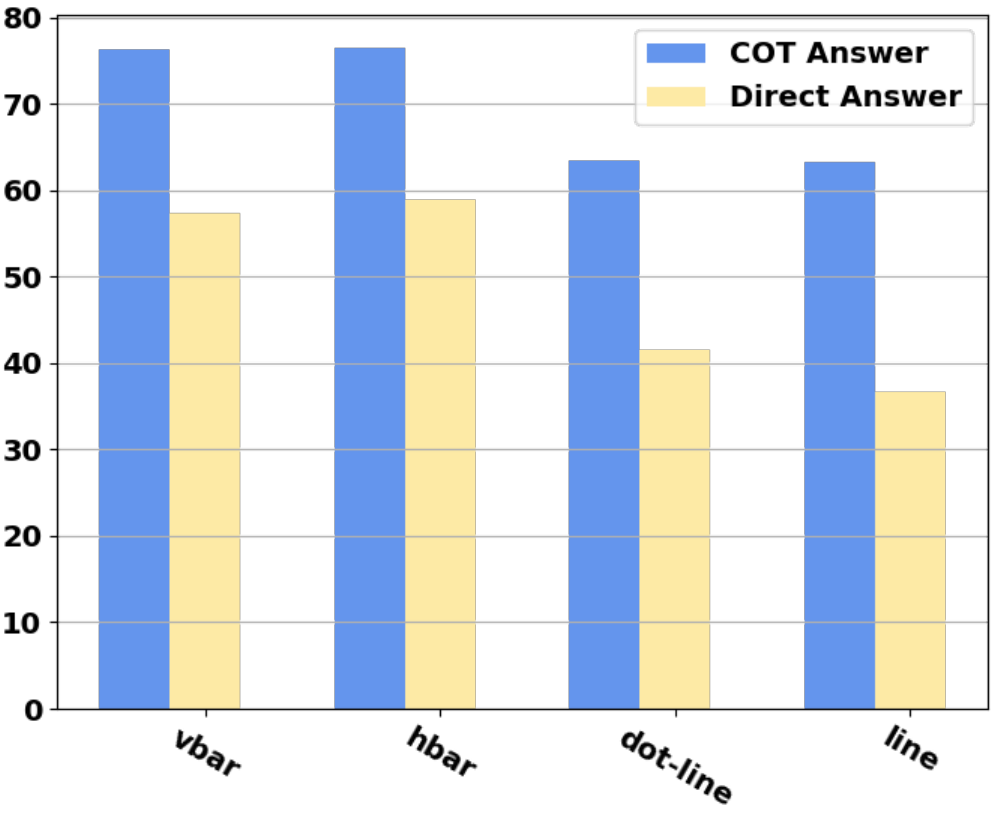} 
\caption{A comparison of the results of using COT answer and direct answer on numerical question answering task, which indicates that using COT answer significantly enhances the model's capability in handling chart numerical question answering tasks with all types. }
\label{fig:mathcot}
\end{figure}

\textbf{COT answer vs. Direct answer for numerical question answering.} In Fig.\ref{fig:mathcot}, we compare using COT answer with direct answer in the same training pipeline for the chart numerical question answering task. Using COT answers instead of direct answers increases the accuracy from 51.9\% to 72.1\%, with improvements across all chart types, especially in dot-line and line charts, where accuracy has increased by 22\% and 26.6\% respectively. This improvement indicates the effectiveness of COT answers in elevating the overall accuracy and performance across various chart types, which reflects that using COT answers teaches the model the reasoning steps and offloads the calculations to the backend system, thus boosting the model's mathematical computation ability.

\textbf{Compared with Unichart after task-specific fine-tuning(except for Chart-to-Text).} We employ the same training strategy and train with the identical model to highlight the effectiveness gains from our data. Following Unichart's lead in multitask instruction tuning, as table \ref{tab:compareunichart} shows, we fine-tune the model on various test datasets (apart from Chart-to-Text, it utilizes fine-tuning during testing), resulting in improvements across different tasks surpassing those of Unichart. It is noteworthy that both Unichart and ChartAst-D are trained using Donut, emphasizing the superiority of ChartSFT.

\begin{table}[!t]
 \centering
     \caption{Compared with Unichart after task-specific fine-tuning.} 
 \scalebox{0.55}{\begin{tabular}{l|cccccc}
  
  \toprule
  
   & \multicolumn{2}{c}{ChartQA} & 
  \multicolumn{2}{c}{Chart-to-Text} & \multicolumn{1}{c}{Chart-to-Table}   \\ 
  \cmidrule(lr){2-3} \cmidrule(lr){4-5} \cmidrule(lr){6-6}
  
   Model & aug. & human & Pew & Statista & ChartQA & OpenCQA \\

  \midrule

Ours-D w/o align & 89.0 & 42.1 
  & 13.7 & 38.3 & 89.5 & 14.3   \\ 

Unichart & 87.8 & 43.9 
  & 12.5 & 38.1 & 91.1 & 14.8 \\ \midrule

  Ours-D w/o align(ft) & \textbf{89.6} & \textbf{44.2}
  & \textbf{13.7} & \textbf{38.3} & \textbf{91.4} & \textbf{14.9} \\ \bottomrule
  \end{tabular}
    }

    \label{tab:compareunichart}
\end{table}


\textbf{The impact of each multitask instruction tuning component.} \label{component} We evaluated the impact of each segment in our multitask instruction tuning by excluding one task at a time during training and noting effects on ChartQA performance. As table \ref{tab:component} shows, any omission led to a performance drop. In particular, chart summarization's contribution is smallest, possibly because ChartQA centers on data extraction and numerical question answering and not overall chart understanding. Furthermore, a significant performance decline when the numerical question answering task is excluded underlines its critical importance for the model.

\begin{table}[!t]
\centering
\caption{ChartAssistant multitask instruction tuning ablations on ChartQA.}
\resizebox{0.5\textwidth}{!}{
\begin{tabular}{lccc}
\hline
\textbf{Model} & \multicolumn{3}{c}{\textbf{ChartQA}} \\
\cmidrule(lr){2-4} 
 & aug. & human & avg. \\
\hline
ChartAst-D  & \textbf{91.3} & \textbf{45.3} & \textbf{68.3} \\
No Chart Summarization & 90.0 & 43.5 & 66.7 \\
No Open-ended Question Answering & 89.5 & 41.1 & 65.3 \\
No Numerical Question Answering & 88.6 & 38.6 & 63.6 \\
No Chart-to-Table Translation & 88.8 & 41.0 & 64.9 \\
No Referring Question Answering & 89.2 & 41.2 & 65.2 \\
\hline
\end{tabular}
}

\label{tab:component}
\vspace{-0.2in}
\end{table}

\textbf{Key components of ChartSFT analysis.} \label{keycomponent} For reasoning tasks involving specific numerical values, such as ChartQA, as shown in table \ref{tab:component}, the math question-answering task benefits greatly from this, especially, as illustrated in fig .\ref{fig:mathcot}, training in COT-format can significantly enhance the accuracy of mathematical computation problems. For tasks involving the output of long texts, such as openCQA, as demonstrated by table \ref{tab:ablationocr} and table \ref{tab:ablationarxiv}, we find that incorporating a question-answering dataset composed of arXiv data can to some extent improve the performance of these tasks. We believe this is due to the broad scope, diversity, and specificity of the arXiv data. Moreover, compared to SciGraphQA \cite{li2023scigraphqa}, the arXiv data we provide has precise numerical values, results in higher quality question generation. Lastly, thanks to the robust language capabilities of GPT-3.5, it is capable of generating high-quality, comprehensive question-answering datasets.

\section{Conclusion}
Our work is aimed at developing a generalized multimodal model for chart-related tasks. We propose ChartSFT, a comprehensive and expansive dataset with the most diverse range of supported chart tasks and types. In conjunction, we suggest ChartAssistant, a multimodal model trained using a two-stage strategy over ChartSFT, which can achieve state-of-the-art results across multiple chart-related downstream tasks. Through detailed experiments, we further demonstrate the superiority of ChartAssistant. 




\newpage
\newpage
\bibliographystyle{ieeenat_fullname}
\bibliography{main}
\newpage
\newpage
\input{X_suppl}






\renewcommand\thesection{\Alph{section}}
\setcounter{section}{0}


\begin{figure*}[tb!]
\centering
\includegraphics[width=0.9\linewidth]{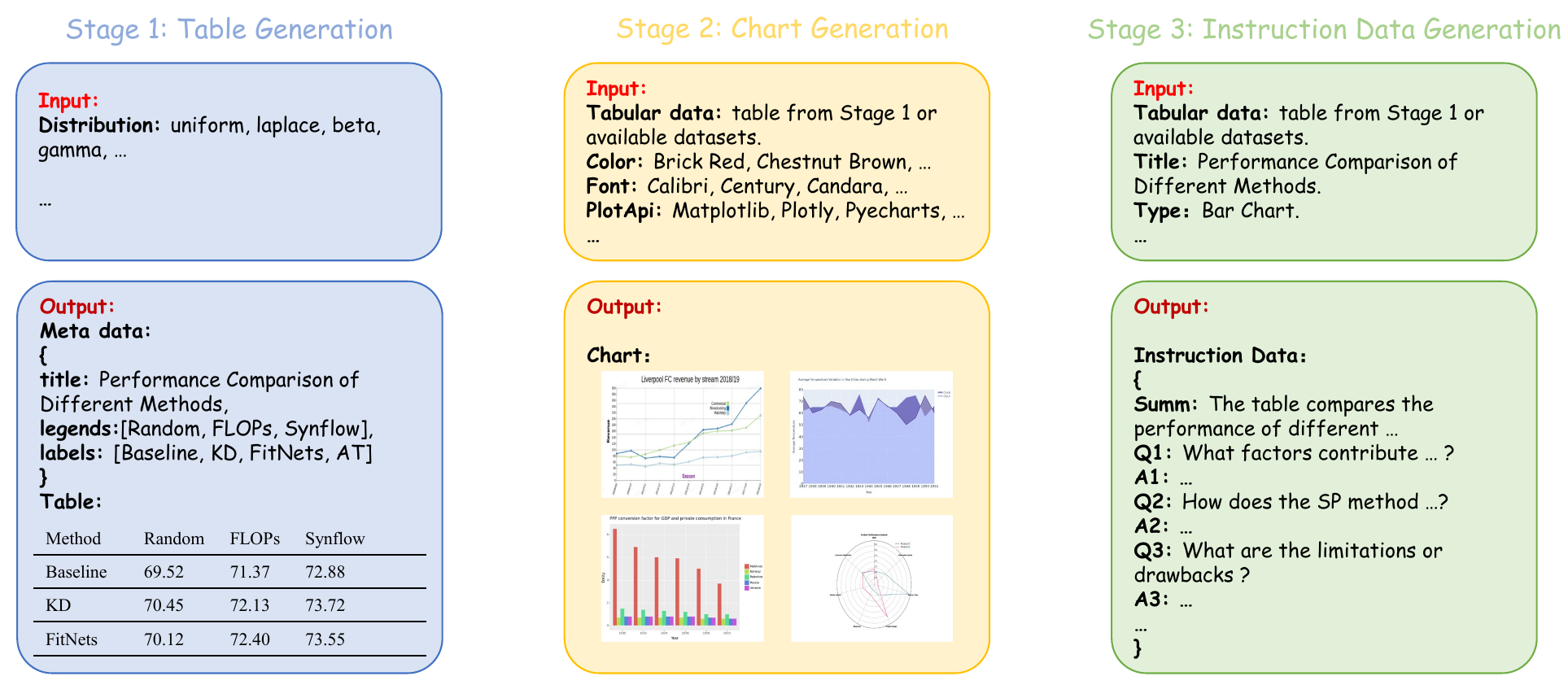} 
\caption{The pipeline of Chart Data Generation in ChartSFT, which consists of three important stages. }
\label{fig:data_gen}
\end{figure*}

\section{ChartSFT}

\subsection{Details of Chart Data Generation in ChartSFT}\label{sec:appen-chartsft-spe}

We illustrate the pipeline of data generation in Fig. \ref{fig:data_gen}. Concretely, the chart data are generated in the following stages:

\textbf{Stage 1: Table generation:} Taking into account the diversity of tabular data, we have predefined over 20 types of probability density distributions, including normal distribution, uniform distribution, beta distribution, Laplace distribution, and more. For each sample, we randomly choose one type of probability density distribution and utilize it to generate values. For different types of charts, we impose further constraints on these values based on their characteristics. (e.g., value range, ratio of positive and negative values, range interval). For radar, bubble and area charts, We directly utilize randomly generated values as the tabular data. For histogram and box plot, we generate an array of extensive values using this distribution and calculate the statistical metrics of this array to serve as the tabular data (e.g., frequencies corresponding to histograms, upper whiskers corresponding to box plots). And then we use the generated data to prompt ChatGPT for creating titles, legends, and labels that align with the numerical characteristics.

\textbf{Stage 2: Chart generation:} To ensure the diversity of the generated charts, we utilize multiple plot APIs, such as matplotlib, plotly, pyecharts, ggplot, seaborn, altair, and more, to plot a variety of styles of the chart. For each chart, we randomly select the following parameters: line (style, thickness), font (style, size, bold, italic), colors, markers, the position of  the elements (title, labels, legends), the size of the charts and so on. Besides our own synthetic tabular data, we also use the table from PlotQA \cite{methani2020plotqa}, ChartQA \cite{masry2022chartqa}, ChartSumm \cite{rahman2022chartsumm} and Chart-To-Text \cite{kantharaj2022chart} to plot the charts for area and radar charts.

\textbf{Stage 3: Instruction Data generation:} For the chart summarization and open-ended QA tasks, we instruct ChatGPT to build datasets by supplying both the table and the corresponding types of charts. For numerical QA and referring QA tasks, we adhere to the approach of the chart with base types by crafting a series of mathematical question templates tailored to the distinct characteristics of various chart types. Subsequently, we manually generate answers with COT annotations.

We adopted a flexible approach by combining ChatGPT with human intervention, which included the utilization of predefined distributions and custom coding of plot API, among other techniques. Through this three-stage chart data generation process, we ensured the diversity and complexity of the table, chart, and instruction data, respectively. As a result, we were able to generate a substantial volume of diversified high-quality chart data.

\subsection{Numerical QA Templates}
We present all the Numerical QA templates in this section. We systematically record both the number of steps in the COT annotation and the number of unique functions used to obtain for each template. Fig \ref{fig:template1} shows 101 general templates designed for charts with different types. However, not all of these general templates are applicable to all types of charts. Hence, we've customized templates to match the unique characteristics of several specific chart types, such as box plots, bubbles, histograms, and pies, as demonstrated in fig . \ref{fig:template6-1}

\subsection{Details of Referring QA in ChartSFT}
In this section, We introduce the details of the generation pipeline of referring QA in our ChartSFT.

\begin{figure*}[htp]
    \centering
    \begin{subfigure}[b]{0.49\textwidth}
        \includegraphics[width=\textwidth]{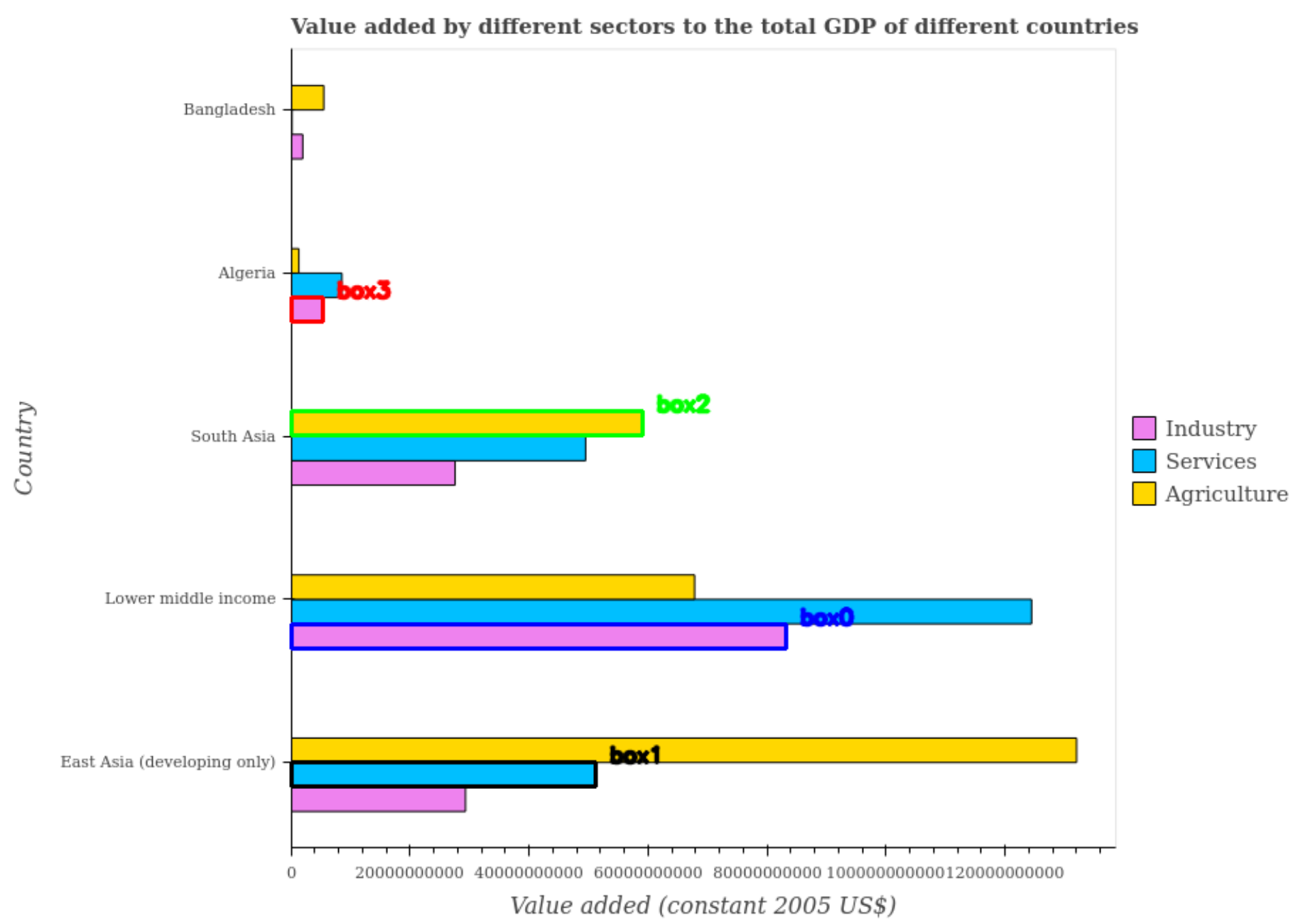}
        \caption{bar chart with referring boxes}
    \end{subfigure}
    \begin{subfigure}[b]{0.49\textwidth}
        \includegraphics[width=\textwidth]{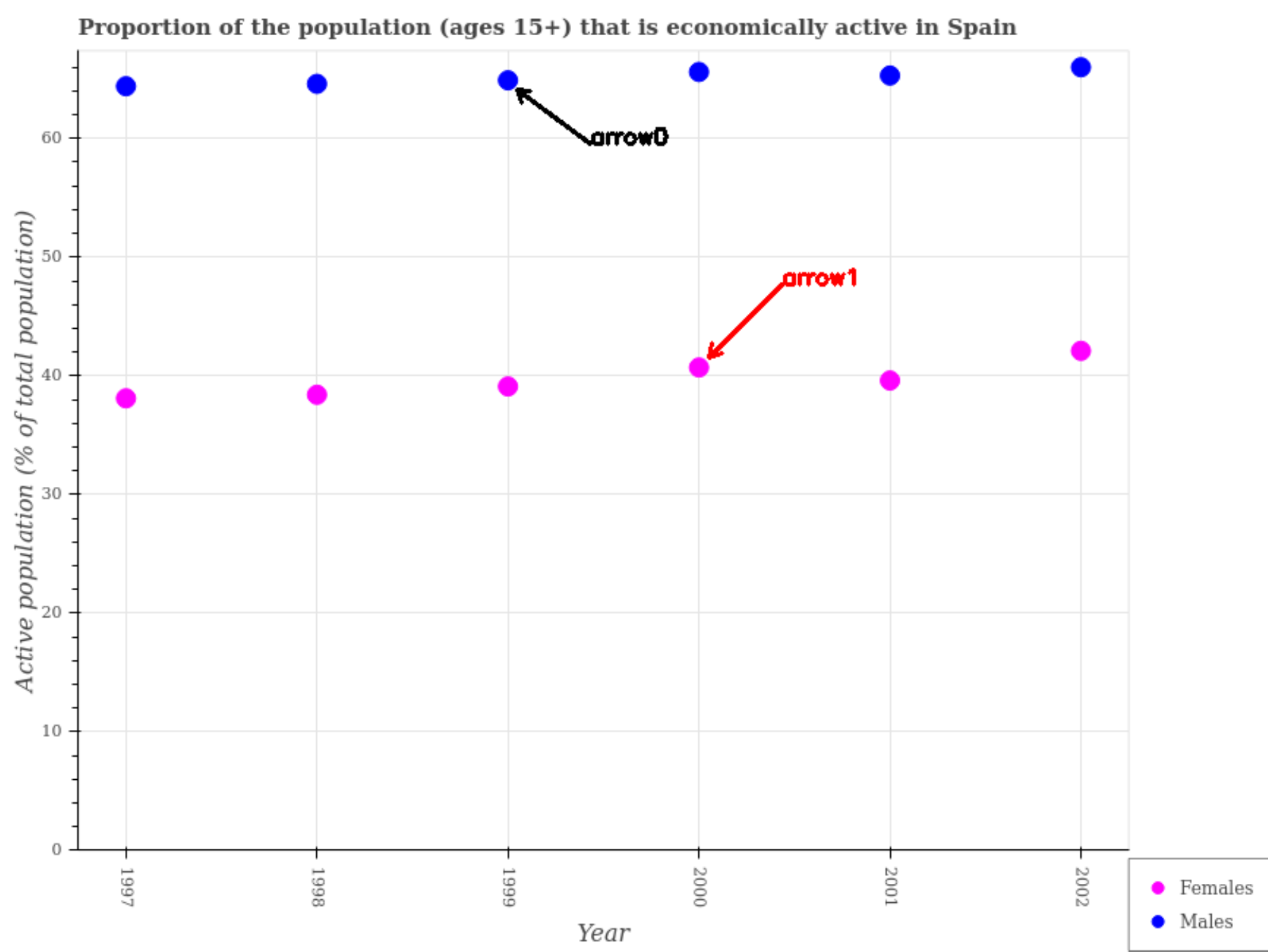}
        \caption{dot-line chart with referring arrows}
    \end{subfigure}
        \begin{subfigure}[b]{0.49\textwidth}
        \includegraphics[width=\textwidth]{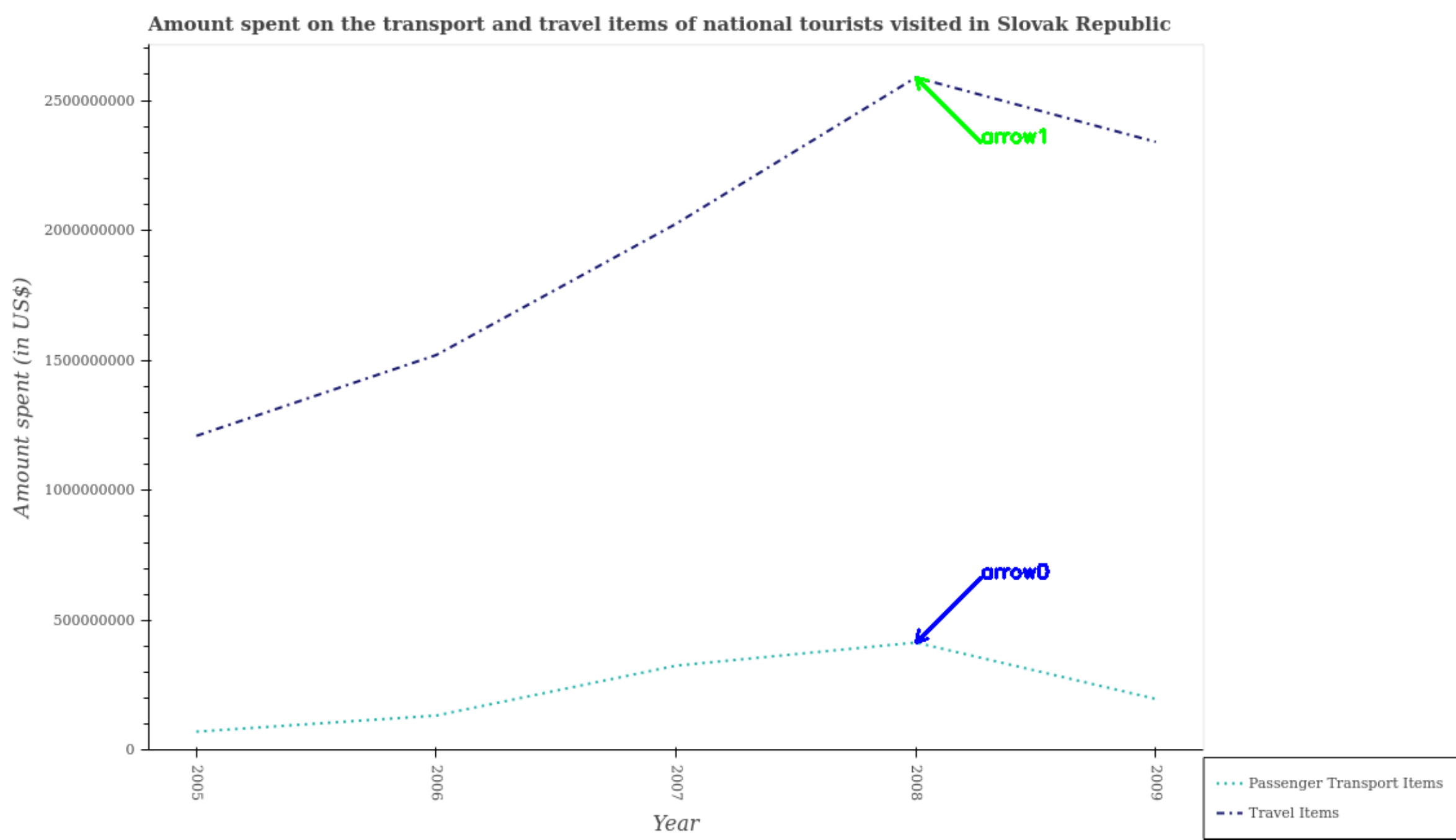}
        \caption{line chart with referring arrows}
    \end{subfigure}
    \begin{subfigure}[b]{0.49\textwidth}
        \includegraphics[width=\textwidth]{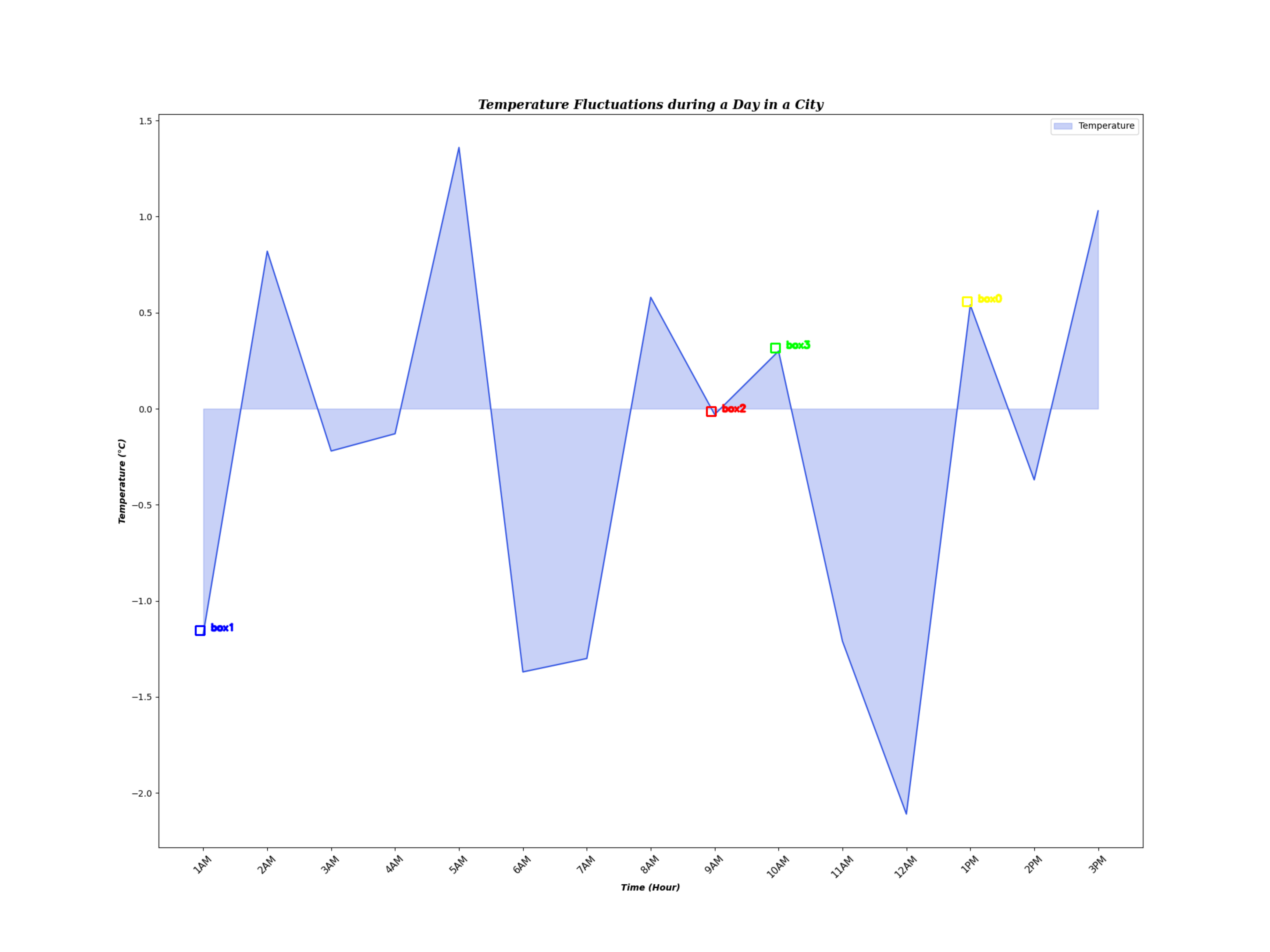}
        \caption{area chart with referring boxes}
    \end{subfigure}
        \begin{subfigure}[b]{0.49\textwidth}
        \includegraphics[width=\textwidth]{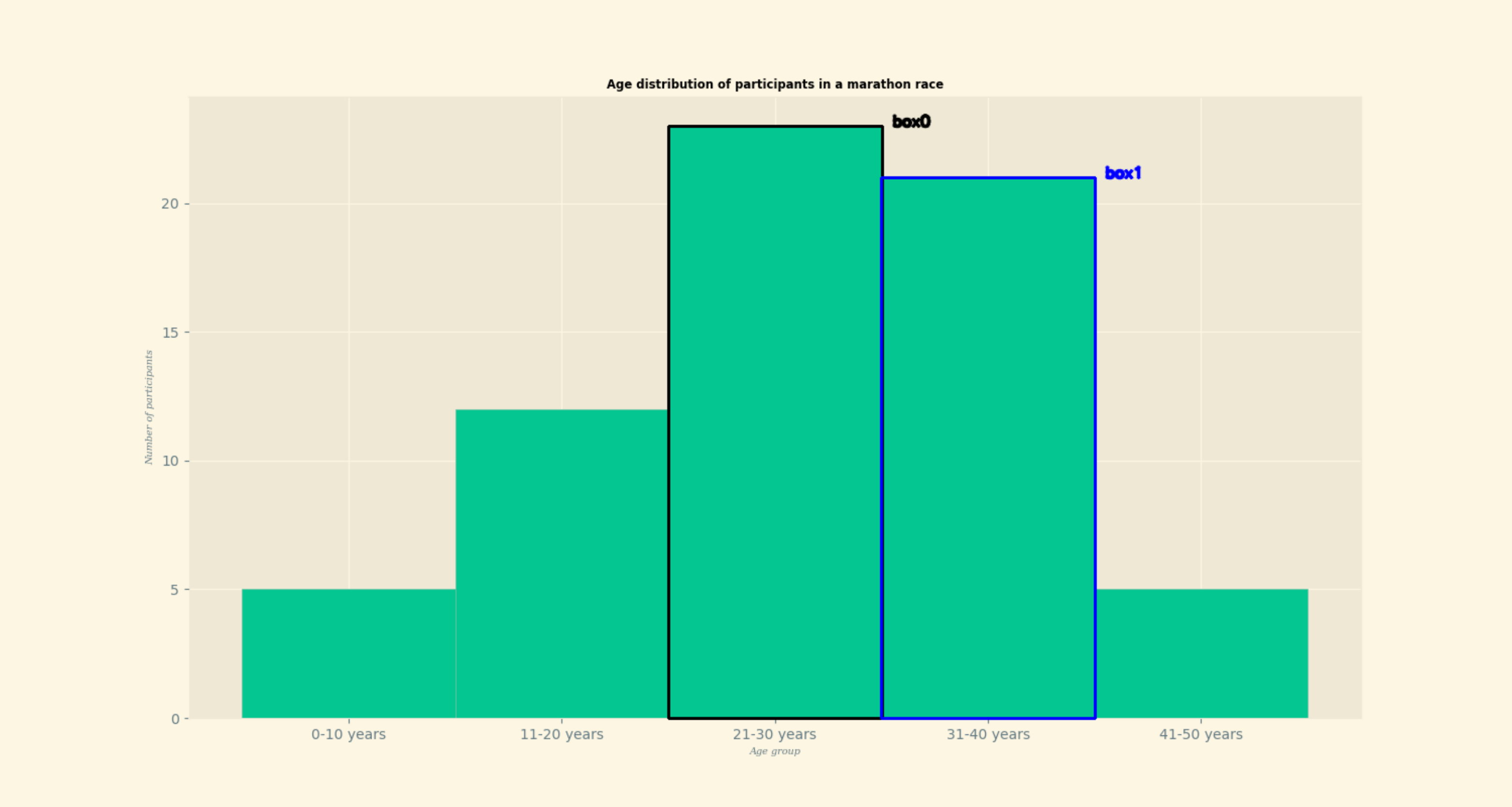}
        \caption{histogram chart with referring boxes}
    \end{subfigure}
        \begin{subfigure}[b]{0.49\textwidth}
        \includegraphics[width=\textwidth]{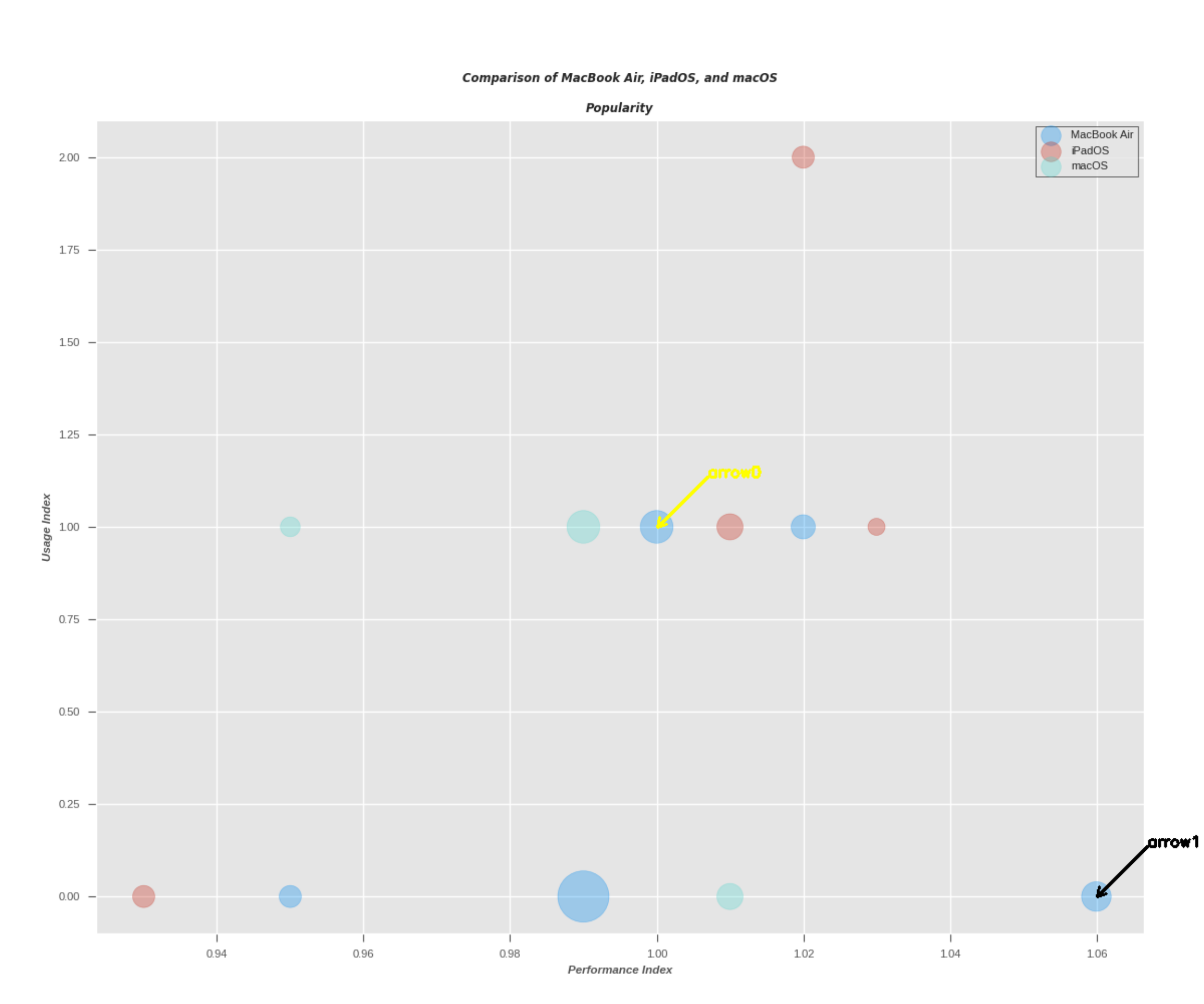}
        \caption{bubble chart with referring arrows}
    \end{subfigure}
    
    \caption{Some examples of different types of charts with referring markers.}
    \label{fig:bbox-example}
\end{figure*}

\textbf{Chart Generation.} We generate charts with the referring box in two ways. 1) For base types of charts, we utilize the bounding box annotations from plotQA to add referring markers onto their original images. 2) For specialized types of charts, we directly generate charts with integrated referring markers leveraging certain Python API(e.g., matplotlib) functionalities. Fig.\ref{fig:bbox-example} shows different types of charts with different referring markers.

\textbf{QA Generation.} Following the pipeline used in generating numerical QA templates, we extend its application to the referring QA task. As outlined in fig. \ref{fig:template6-2}, we define a total of 114 templates, encompassing questions related to label recognition and mathematical calculations. Note that the x\_tick of line and area charts is continuous, therefore, we tailor these templates to accommodate such scenarios.

\begin{figure}[tb!]
\centering
\includegraphics[width=0.8\linewidth]{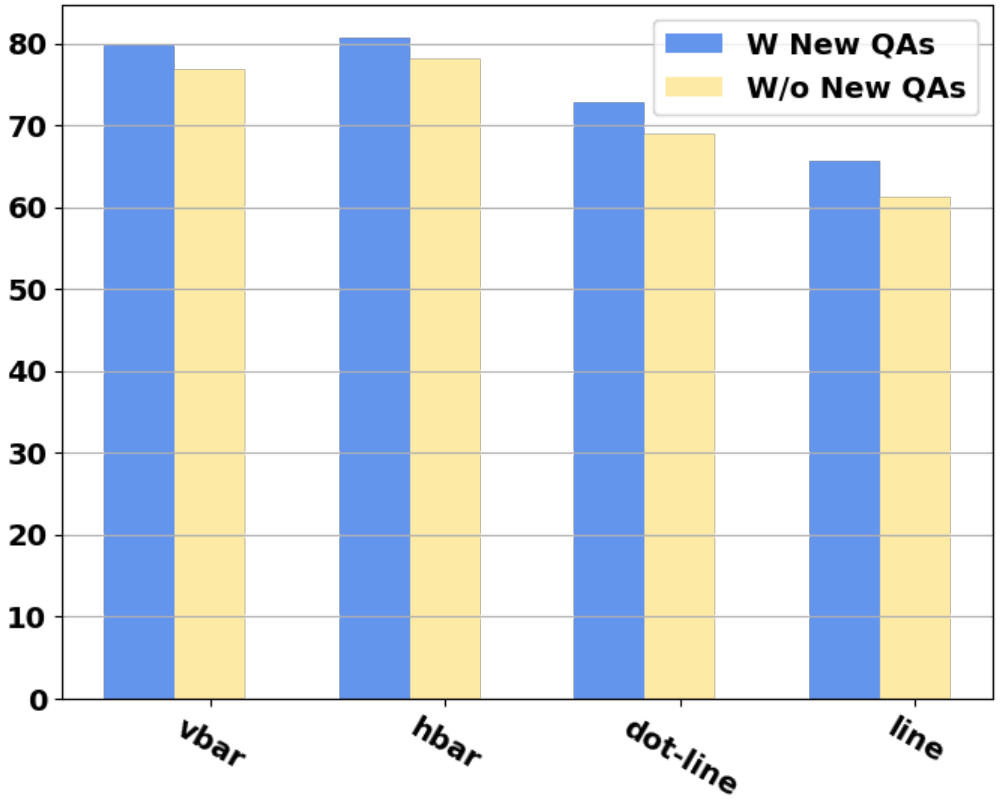} 
\caption{A comparison of the results of training with new question-answer pairs or not, which indicates that incorporating equivalent questions into the training process can enhance the model's robustness towards math questions. }
\label{fig:newqa}
\end{figure}

\section{Experiments}

\subsection{Experimental Setups}\label{ExperimentalSetups}
We begin by conducting alignment pre-training, utilizing the chart-to-table translation task for 65k steps. Following that, we engage in multitask instruction tuning. We employ the Adam optimizer \cite{kingma2014adam} with a scheduled learning rate, where the initial rate is set to 5e-5 for ChartAst-D and 2e-6 for ChartAst-S. The input resolution is established at 224×224 and 448×448, while the maximum length in the decoder is defined as 1536 for ChartAst-D and 2048 for ChartAst-S. After training for four epochs for ChartAst-D and only one epoch for ChartAst-S, we perform testing on multiple downstream tasks. During inference, each task receives an image and a textual instruction as input, and the model generates a textual answer. All training processes are carried out on 16xA100 80GB GPUs. ChartAst-S outperforms ChartAst-D and has stronger robustness. This is partly due to the special high-resolution image handling method employed by ChartAst-S, which retains more detailed chart information. Additionally, ChartAst-S incorporates richer pre-training knowledge and the larger model possesses greater robustness.

\textbf{Evaluation.}  We assess the performance of ChartAssistant across various tasks and datasets. Following the evaluation of Unichart \cite{masry2023unichart}, we utilize Chart-to-text \cite{kantharaj2022chart} for evaluating chart summarization task,  and OpenCQA \cite{kantharaj2022opencqa} and ChartQA \cite{masry2022chartqa} for open-ended question answering task. The ChartQA dataset consists of two subsets: augmented and human. The augmented set comprises machine-generated summaries with a predominantly extractive nature, while the human set contains manually crafted summaries that require more advanced reasoning. The Chart-to-Text task encompasses two sets named "Pew" and "Statista" indicating the origin of the image examples. In the Pew set, summaries are automatically extracted from areas surrounding the images, while in the Statista set, summaries are authored by human annotators. We use ChartQA and PlotQA to evaluate chart-to-table translation tasks due to their various chart styles. To evaluate numerical question answering and referring question answering, we sample test sets from the datasets constructed by ourselves called MathQA and ReferQA.

\textbf{Metrics.} For evaluating ChartQA, MathQA, and ReferQA, we adopt relaxed correctness, which allows for an exact match with tolerance for a 5\% numerical error \cite{liu2022matcha,masry2023unichart}. As for Chart-to-Text and OpenCQA, we employ BLEU as the evaluation metric following previous works \cite{liu2022matcha,masry2023unichart}. For chart-to-table translation, we use $RMS_{F1}$ from DePlot \cite{liu2022deplot}. This metric is resilient to modifications such as transpositions or permutations of columns and rows and has the capacity to accommodate and impose penalties for minor errors in numerical or textual data up to a specified threshold. At the same time, it can distinctively illustrate any reductions in both precision and recall. To cater for table transpositions, we evaluate both the original table and its transposed version and select the highest $RMS_{F1}$ score.

\begin{table}[!t]
\centering
\caption{The chart-to-table translation performance of ChartAssistant and some baselines on plotQA.}
\resizebox{0.40\textwidth}{!}{
\begin{tabular}{lcccc}
\hline
Dataset & ChartAst-S & ChartAst-D & Matcha & Unichart \\
\hline
PlotQA  & \textbf{95.6} & 90.1 & 82.7 & 70.8 \\

\hline
\end{tabular}
}

\label{tab:plotqa-ocr}
\end{table}

\begin{figure*}[tb!]
\centering
\includegraphics[width=0.98\linewidth]{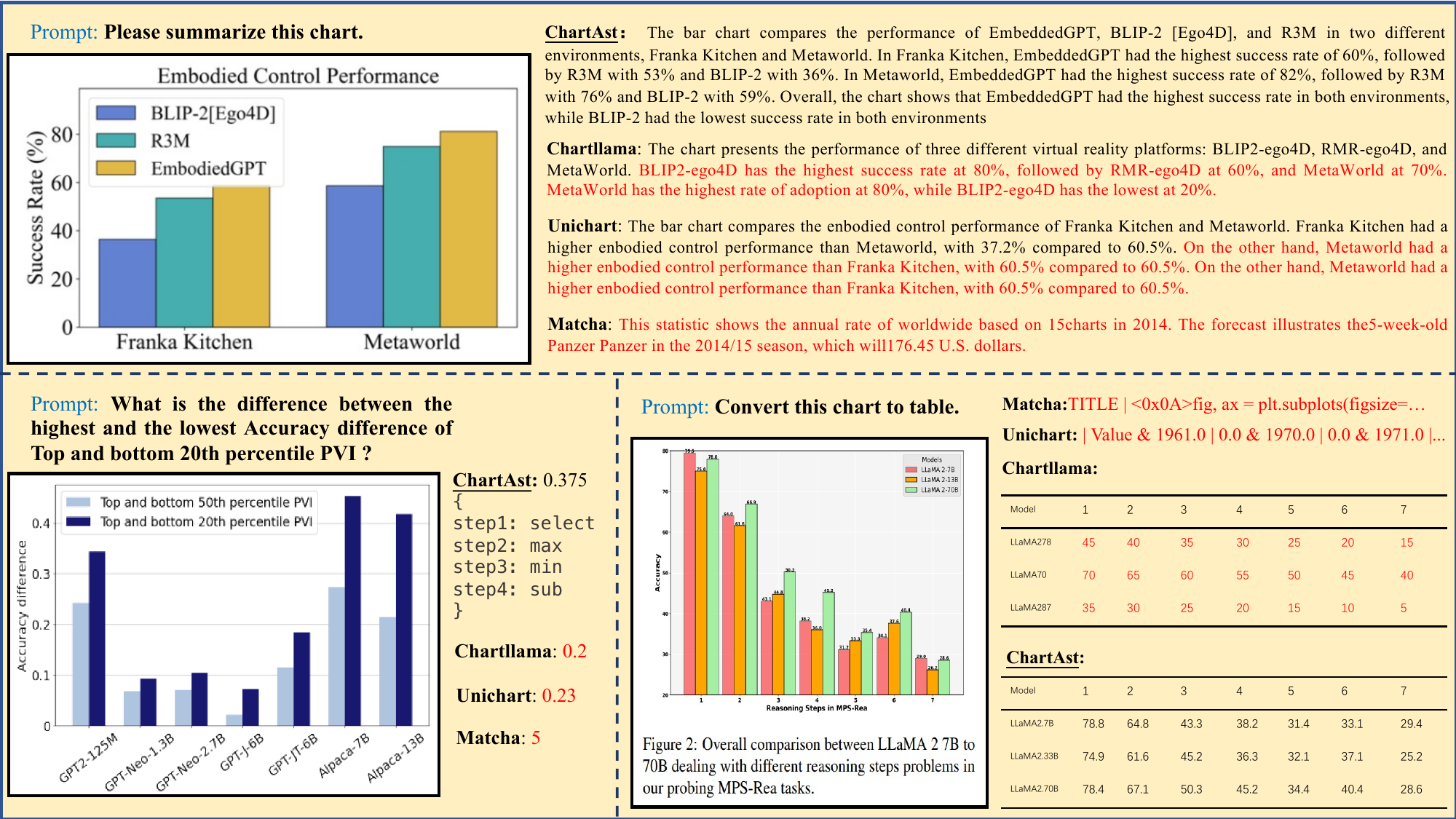} 
\caption{ChartAst-S demonstrates outstanding generalization ability in chart-to-table translation, summarization, and question-answering tasks.} 
\label{fig:zero-shot-demo}
\end{figure*}

\textbf{More experiments.} A significant portion of the ChartQA \cite{masry2022chartqa} dataset labels corresponding numerical data on the charts, but there also exists a considerable amount of charts where the numbers are not visualized. Consequently, we utilize the PlotQA \cite{methani2020plotqa} dataset to conduct additional chart-to-table translation experiments. As table \ref{tab:plotqa-ocr} shows, the results indicate that compared to the ChartQA dataset, the ChartAssistant demonstrates a more significant advantage when implemented on the PlotQA dataset.

\section{Ablation Study}

We thoroughly analyze the key aspects of our approach. In the appendix, we consider the significance of arXiv data and the impact of generating equivalent math questions on the effectiveness and robustness of our approach.

\textbf{The impact of generating equivalent math questions.} Considering that generating questions purely through templates can be rather rigid in the math question answering task, we attempt to provide both the template questions and table information to ChatGPT simultaneously, asking it to generate more significant equivalent questions based on the meaning of the tables. In particular, "What is the difference between the highest and the lowest Amount of Least developed countries ?" can be converted to "What is the range of the Amount for Least developed countries ?". We divide these new question-answer pairs into training and test sets, then compare the performance on the test set when training with and without this additional data.

As fig. \ref{fig:newqa} demonstrates, we find that including the newly generated equivalent questions in the training can enhance the performance of all types compared to the original approach. In detail, the overall accuracy changes from 71.8\% to 76.2\%.

\section{Some demos from Out of Distribution}

\begin{figure*}[tb!]
\centering
\includegraphics[width=0.98\linewidth]{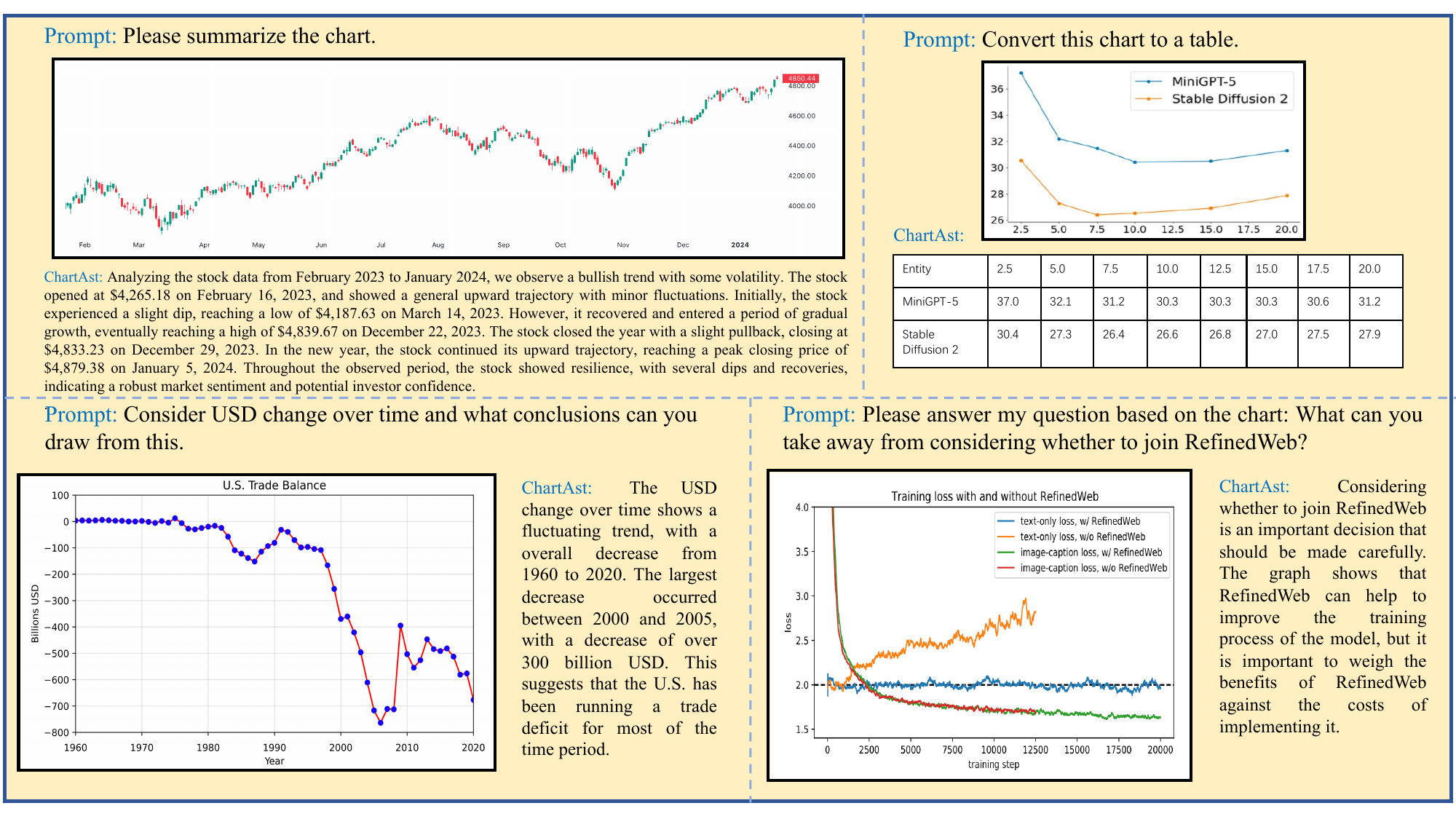} 
\caption{ChartAst-S demonstrates outstanding generalization ability in chart-to-table translation, summarization, and question-answering tasks.} 
\label{fig:demo1}
\end{figure*}

\begin{figure*}[tb!]
\centering
\includegraphics[width=0.98\linewidth]{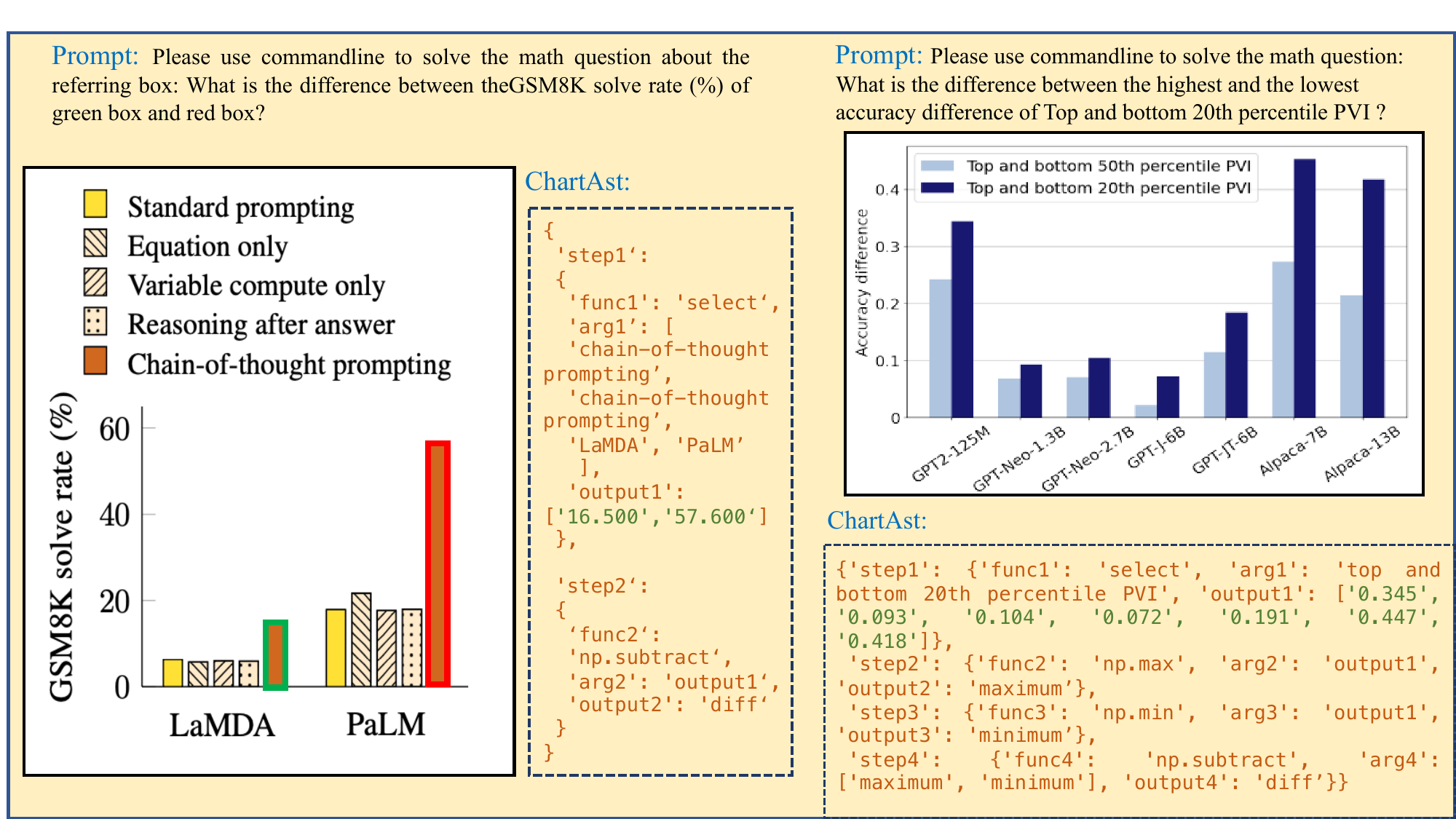} 
\caption{ChartAst-S demonstrates outstanding generalization ability in mathematical and referring question-answering tasks.} 
\label{fig:demo2}
\end{figure*}

To demonstrate the model's generalization capability, we randomly take screenshots of several charts, as shown in Fig .\ref{fig:demo1} and Fig .\ref{fig:demo2} . We find that the model possesses generalization ability on out-of-distribution samples. Additionally, as shown in fig. \ref{fig:zero-shot-demo}, we visualize some demos comparing the performance of zero-shot scenarios with baseline methods. We observe that in summarization tasks, UniChart and Matcha tend to produce repetitions or hallucinations, whereas ChartLlama and ChartAssistant exhibit relatively stronger capabilities in handling summarization tasks. However, ChartLlama commits some factual errors; in question answering, thanks to the incorporation of COT-format QA training data, ChartAssistant effectively addresses QA tasks requiring mathematical reasoning. Lastly, in chart-to-table translation, UniChart and Matcha accurately model the table structure. Although ChartLlama can model the table structure accurately, the values are completely incorrect. Only ChartAssistant successfully constructs the table of the chart accurately.

\clearpage

\begin{figure*}[tb!]
\centering
\includegraphics[width=0.99\linewidth]{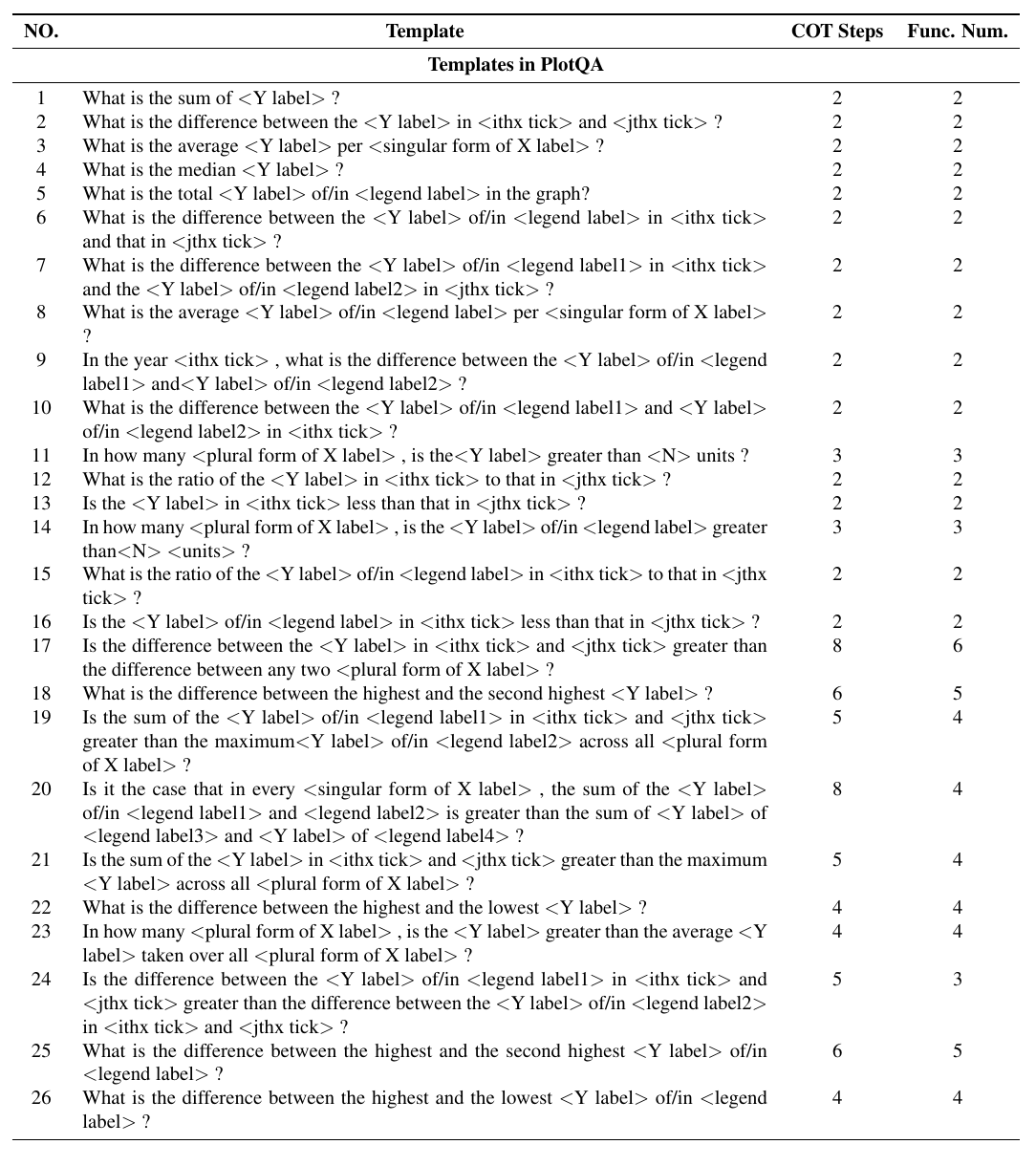} 
\caption{General Numerical QA Templates in ChartBench. Containing 40 template questions from PlotQA and 61 template questions that we designed additionally.}
\label{fig:template1}
\end{figure*}

\clearpage

\begin{figure*}[tb!]
\centering
\includegraphics[width=0.99\linewidth]{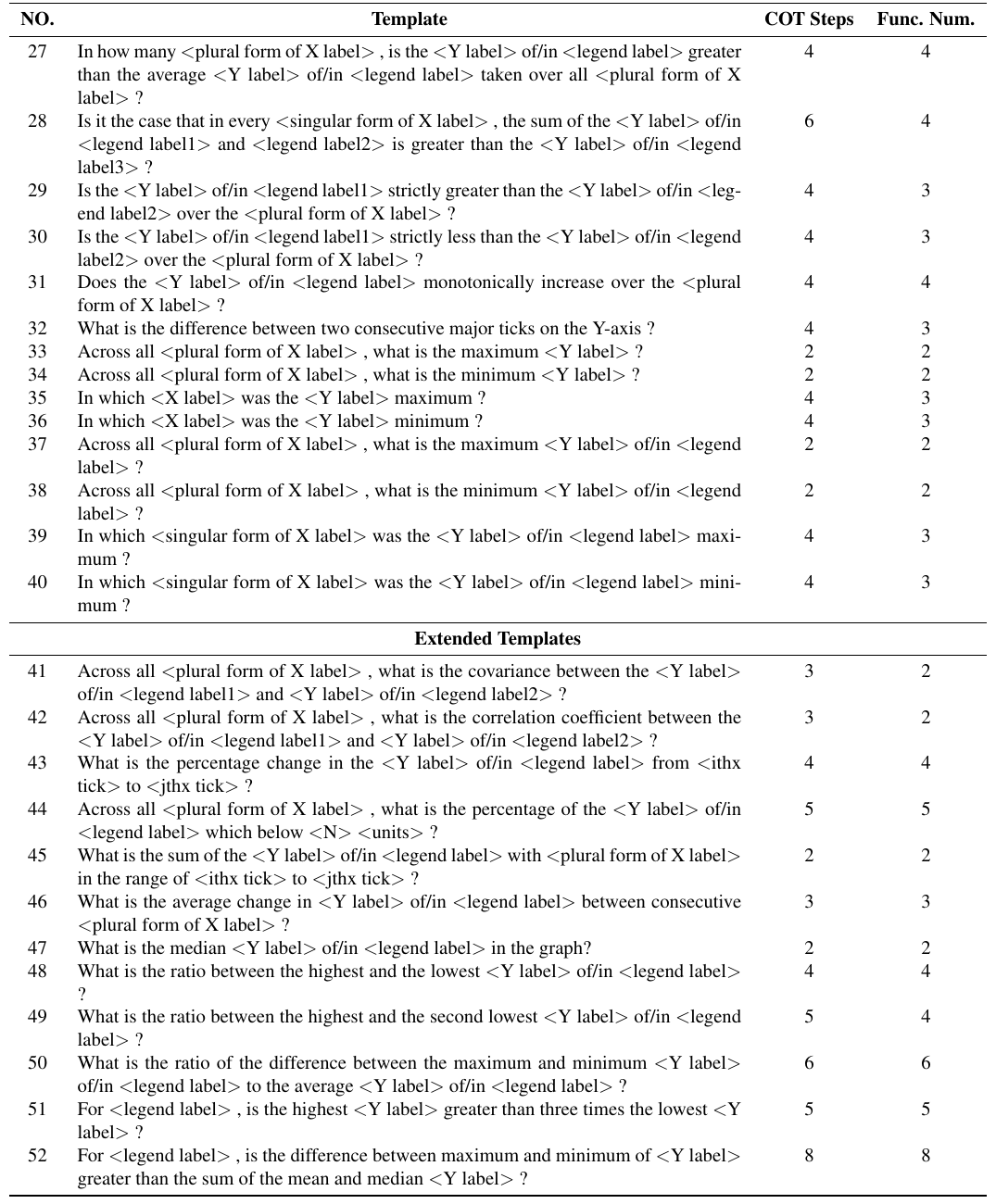} 
\caption{-- continued from previous page.}
\label{fig:template2}
\end{figure*}

\clearpage

\begin{figure*}[tb!]
\centering
\includegraphics[width=0.99\linewidth]{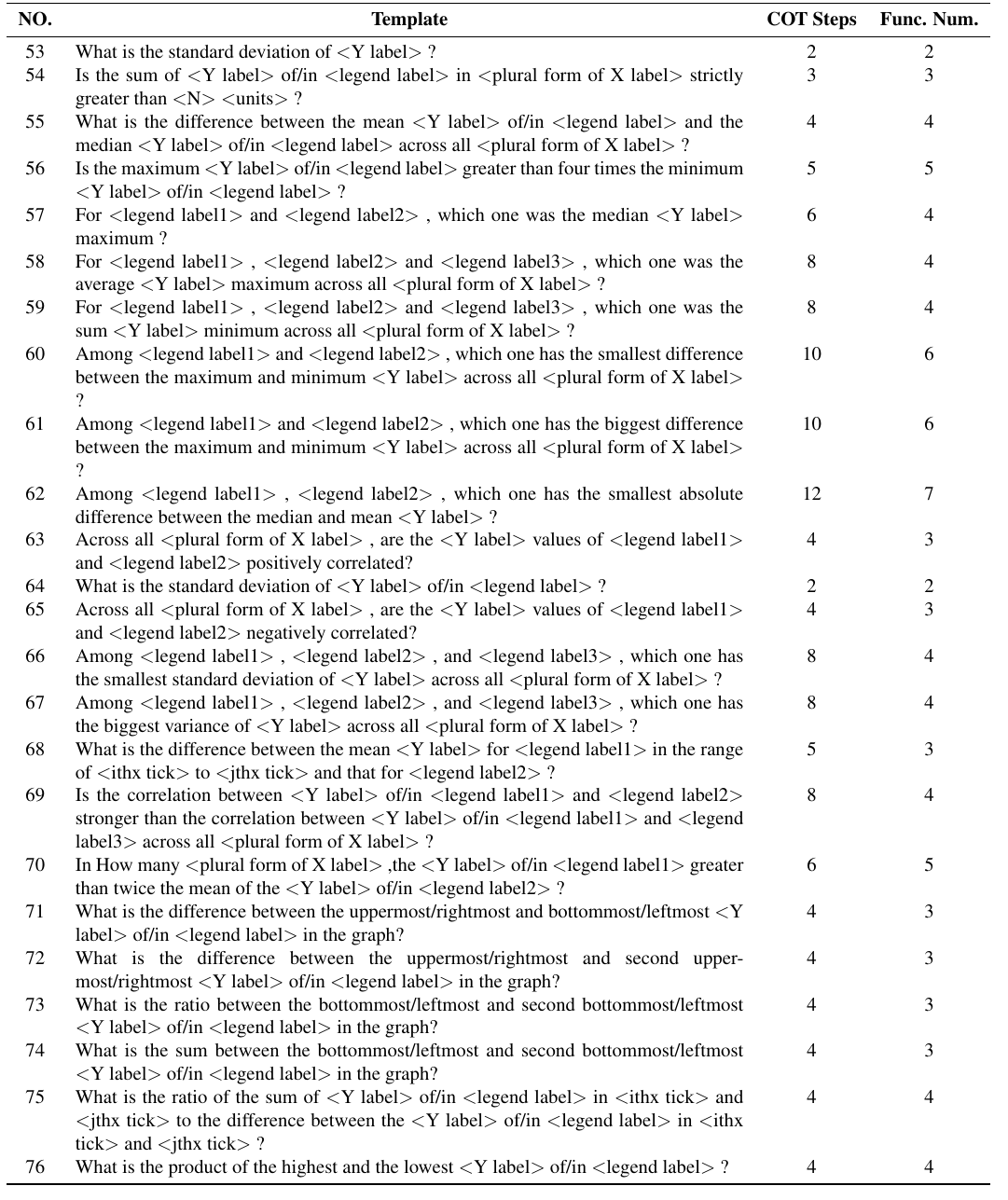} 
\caption{-- continued from previous page.}
\label{fig:template3}
\end{figure*}

\clearpage

\begin{figure*}[tb!]
\centering
\includegraphics[width=0.99\linewidth]{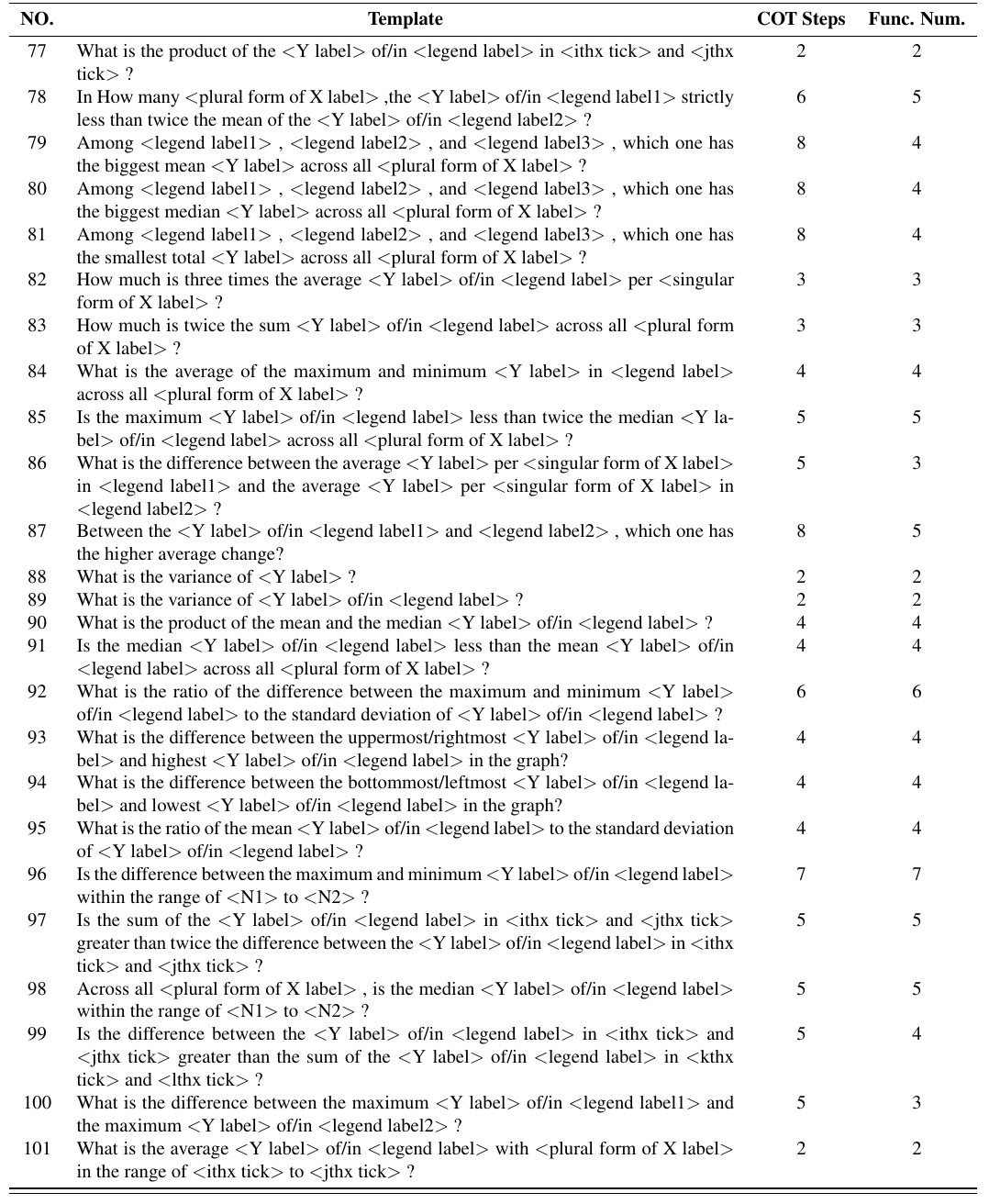} 
\caption{-- continued from previous page.}
\label{fig:template4}
\end{figure*}

\clearpage

\begin{figure*}[tb!]
\centering
\includegraphics[width=0.99\linewidth]{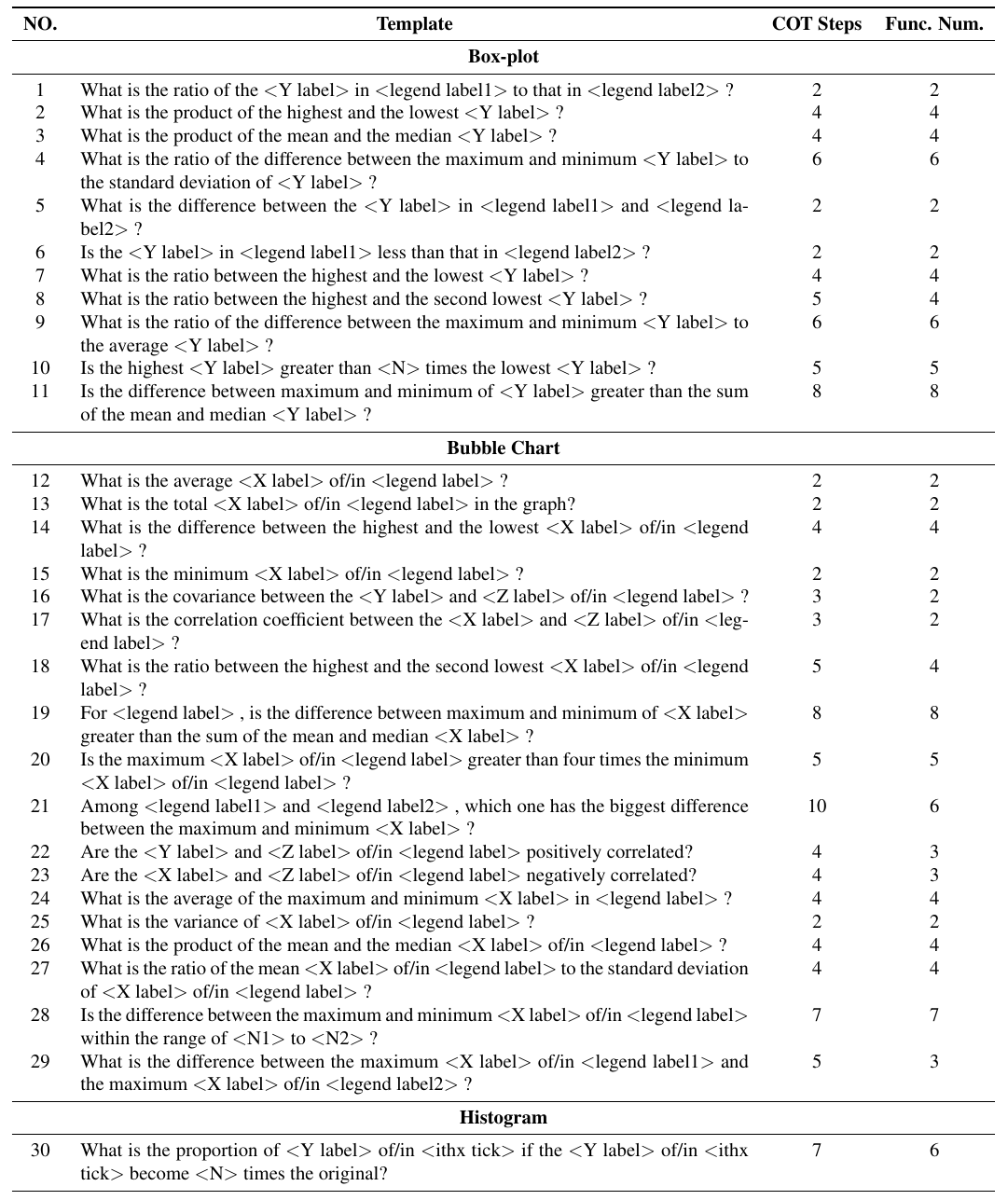} 
\caption{Numerical QA Templates for several Types of charts in ChartBench.}
\label{fig:template5}
\end{figure*}

\clearpage

\begin{figure*}[tb!]
\centering
\includegraphics[width=0.99\linewidth]{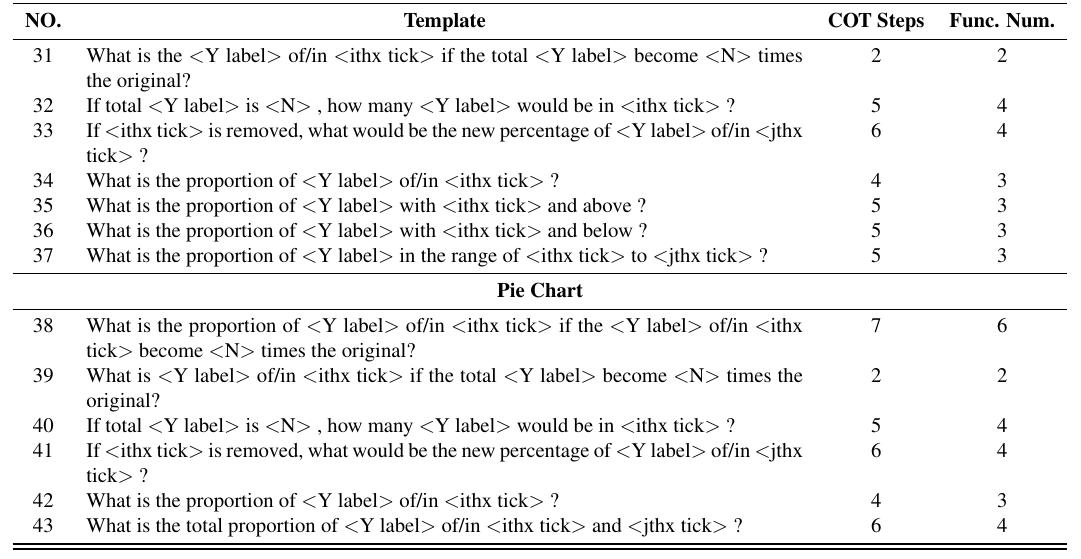} 
\caption{-- continued from previous page.}
\label{fig:template6-1}
\end{figure*}

\begin{figure*}[tb!]
\centering
\includegraphics[width=0.99\linewidth]{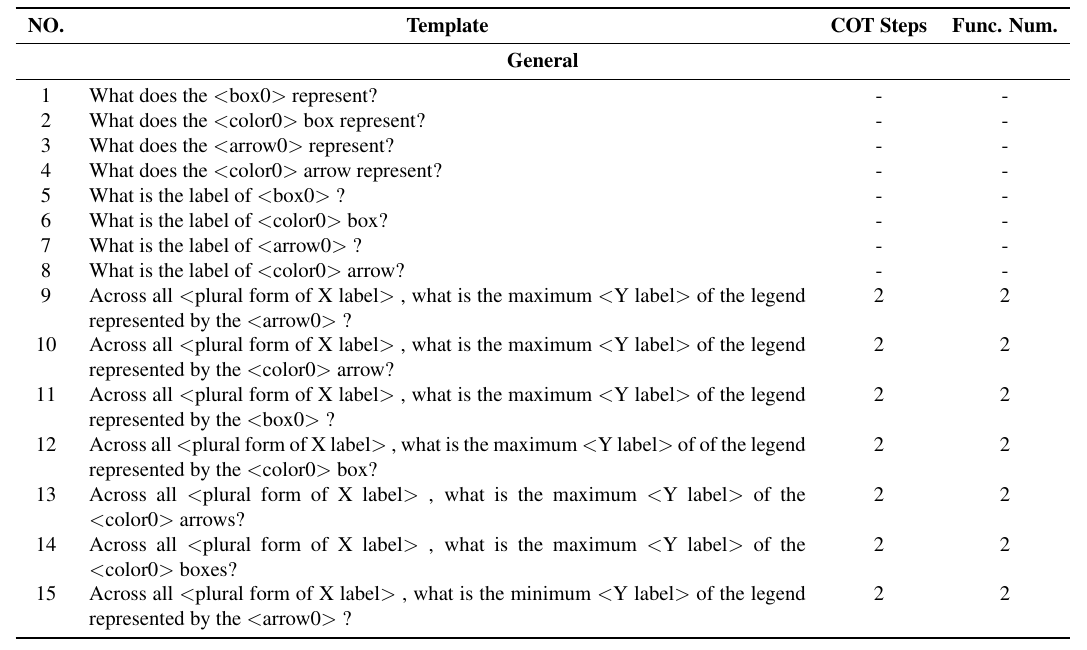} 
\caption{Referring QA Templates in ChartBench.}
\label{fig:template6-2}
\end{figure*}

\clearpage

\begin{figure*}[tb!]
\centering
\includegraphics[width=0.99\linewidth]{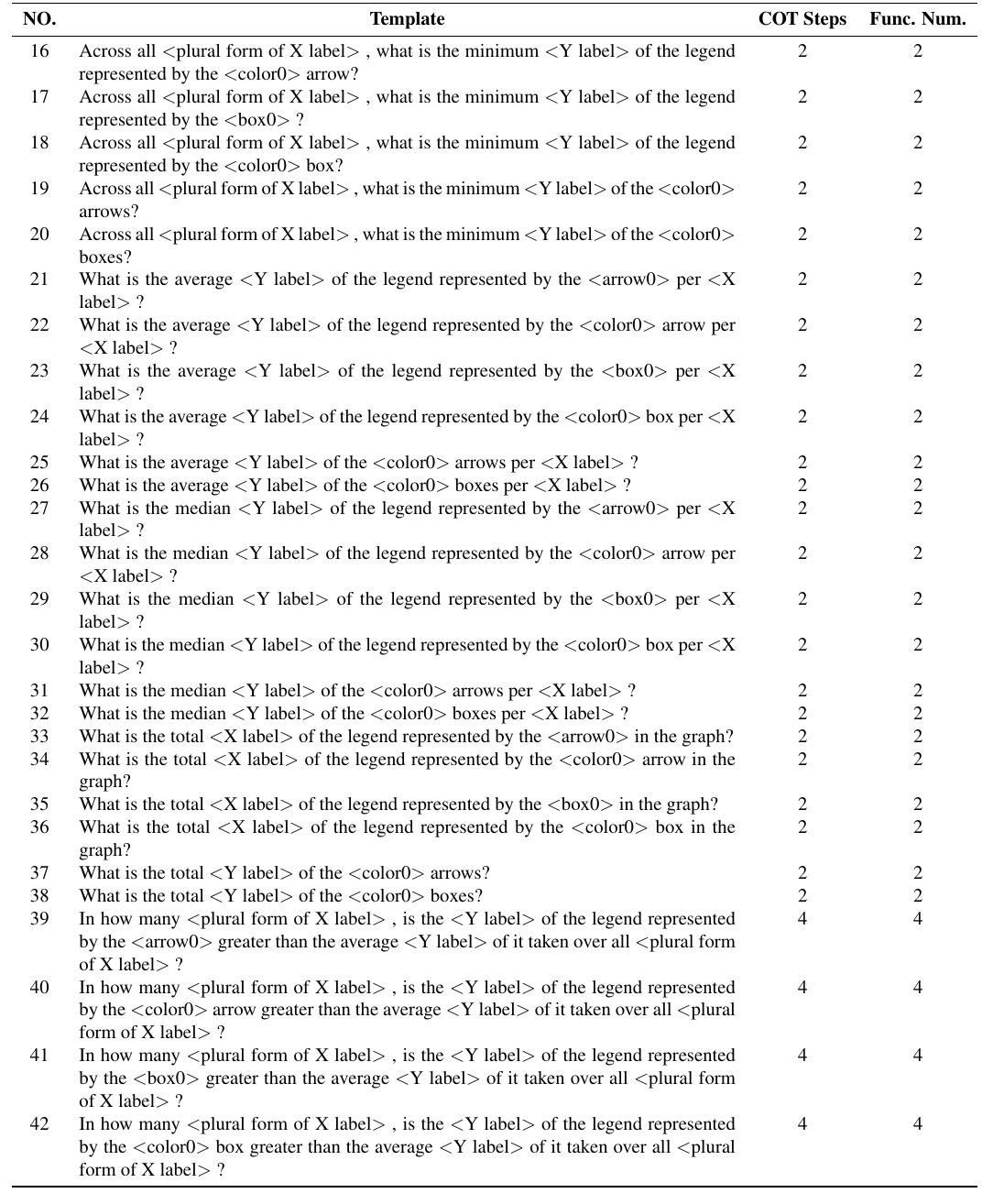} 
\caption{-- continued from previous page.}
\label{fig:template7}
\end{figure*}

\clearpage

\begin{figure*}[tb!]
\centering
\includegraphics[width=0.99\linewidth]{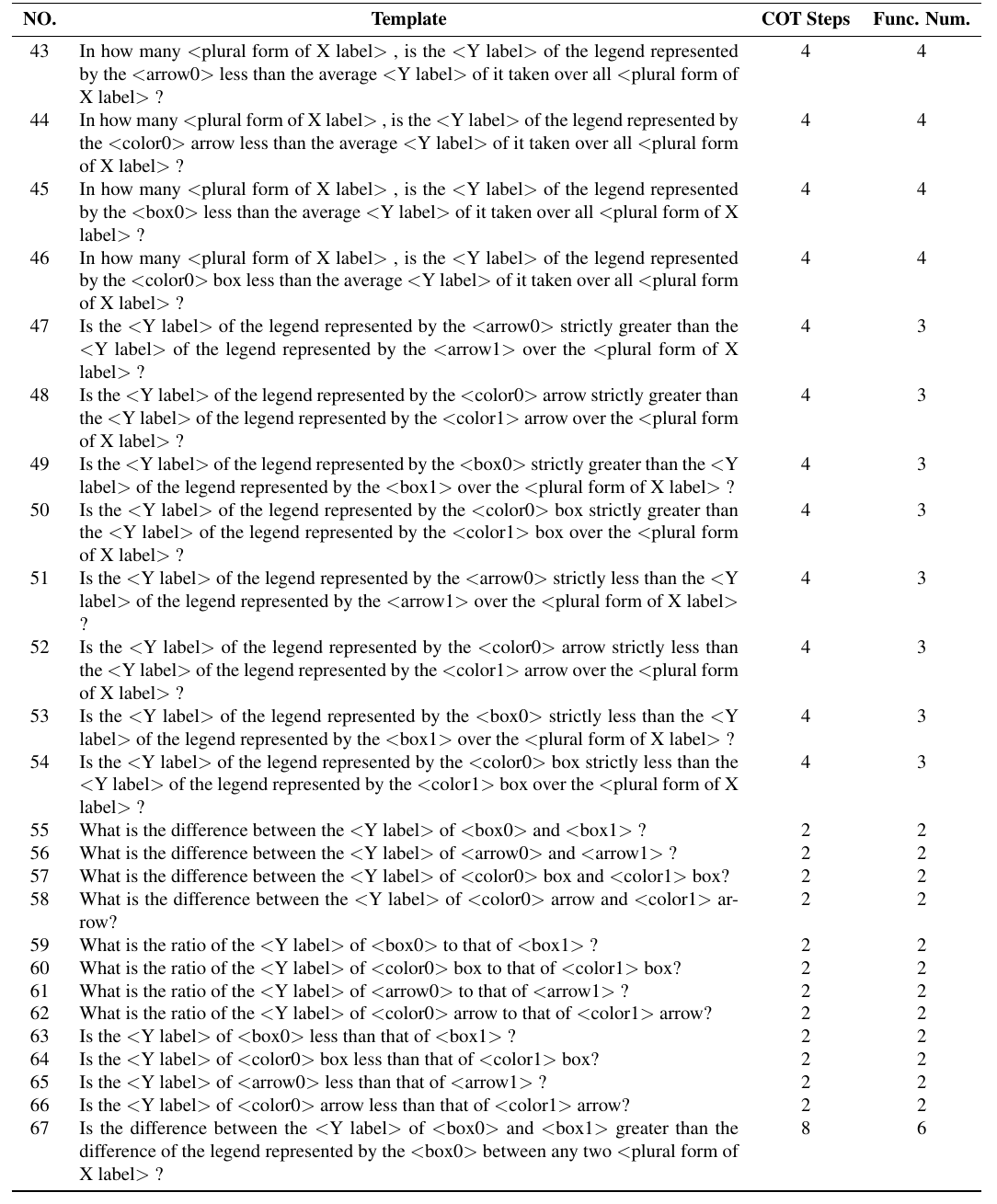} 
\caption{-- continued from previous page.}
\label{fig:template8}
\end{figure*}

\clearpage

\begin{figure*}[tb!]
\centering
\includegraphics[width=0.99\linewidth]{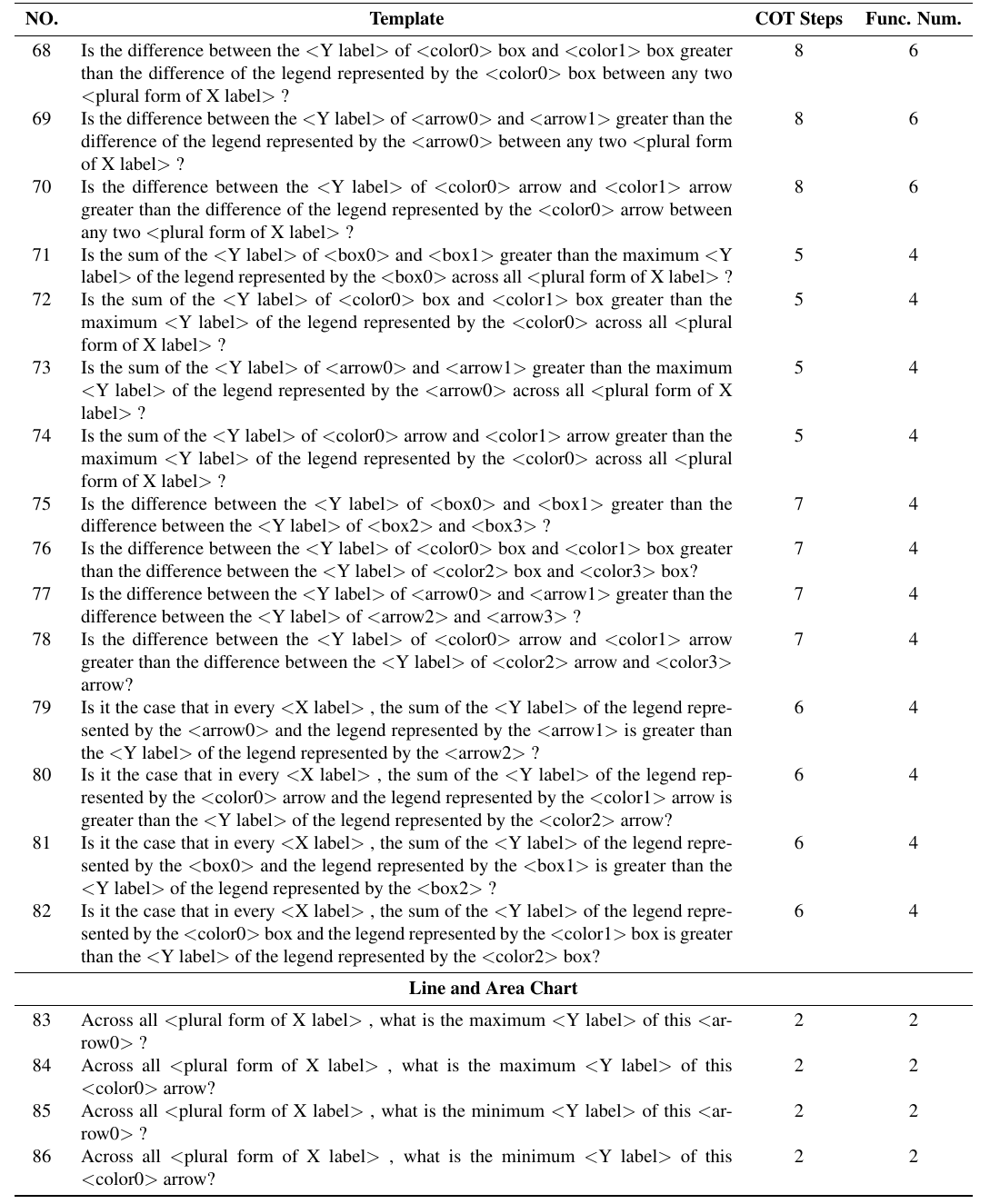} 
\caption{-- continued from previous page.}
\label{fig:template9}
\end{figure*}

\clearpage

\begin{figure*}[tb!]
\centering
\includegraphics[width=0.99\linewidth]{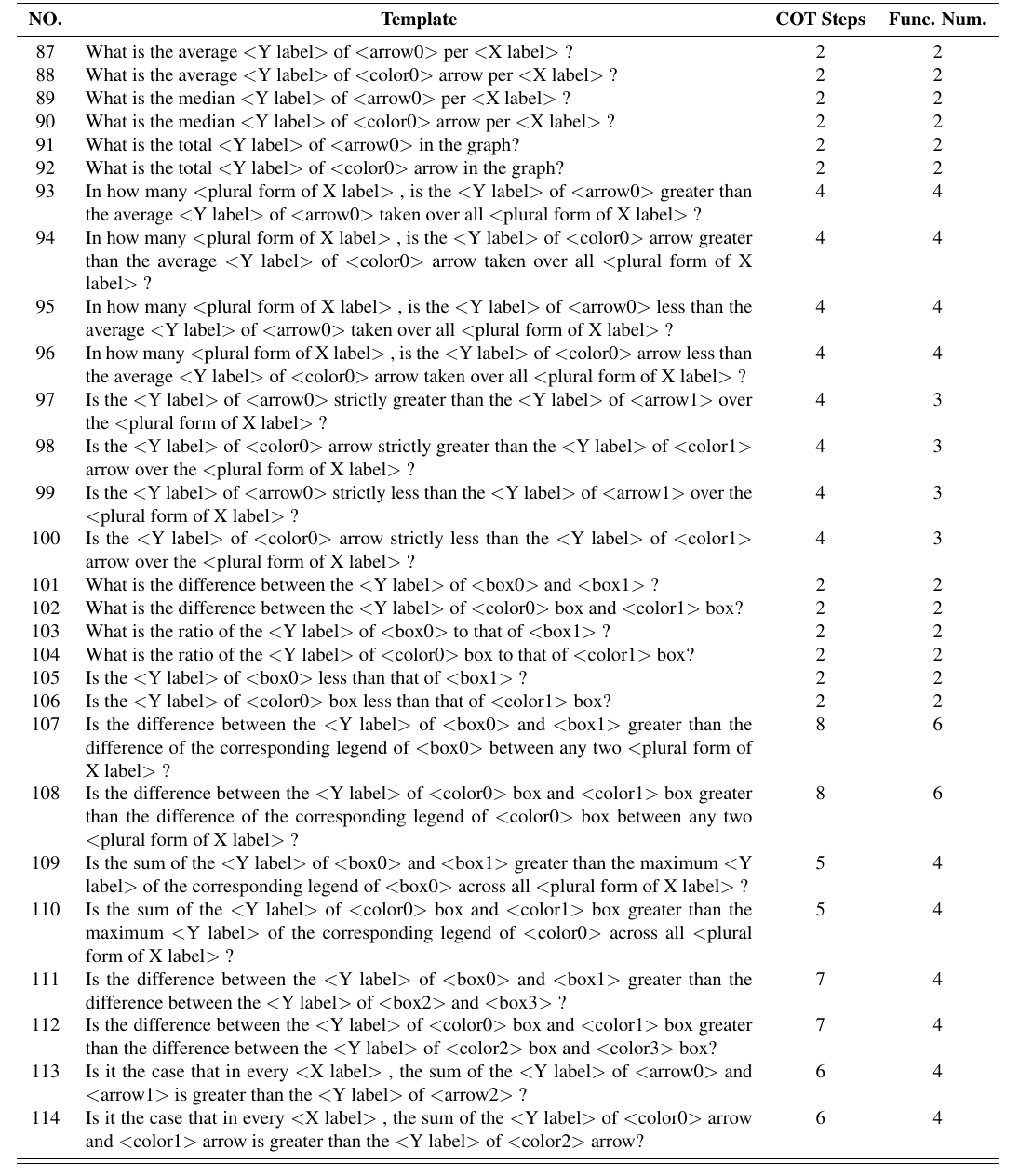} 
\caption{-- continued from previous page.}
\label{fig:template10}
\end{figure*}

\clearpage

\end{document}

%% file: preamble.tex
\usepackage[dvipsnames]{xcolor}


%% file: X_suppl.tex
\clearpage
\setcounter{page}{1}
\maketitlesupplementary


%% file: main.bbl
\begin{thebibliography}{51}
\providecommand{\natexlab}[1]{#1}
\providecommand{\url}[1]{\texttt{#1}}
\expandafter\ifx\csname urlstyle\endcsname\relax
  \providecommand{\doi}[1]{doi: #1}\else
  \providecommand{\doi}{doi: \begingroup \urlstyle{rm}\Url}\fi

\bibitem[arX()]{arXiv}
Arxiv.
\newblock \url{https://arxiv.org/}.

\bibitem[Ahmed et~al.(2023)Ahmed, Jawade, Pandey, Setlur, and Govindaraju]{ahmed2023realcqa}
Saleem Ahmed, Bhavin Jawade, Shubham Pandey, Srirangaraj Setlur, and Venu Govindaraju.
\newblock Realcqa: Scientific chart question answering as a test-bed for first-order logic.
\newblock In \emph{International Conference on Document Analysis and Recognition}, pages 66--83. Springer, 2023.

\bibitem[Bai et~al.(2023)Bai, Bai, Yang, Wang, Tan, Wang, Lin, Zhou, and Zhou]{bai2023qwen}
Jinze Bai, Shuai Bai, Shusheng Yang, Shijie Wang, Sinan Tan, Peng Wang, Junyang Lin, Chang Zhou, and Jingren Zhou.
\newblock Qwen-vl: A frontier large vision-language model with versatile abilities.
\newblock \emph{arXiv preprint arXiv:2308.12966}, 2023.

\bibitem[Blecher et~al.(2023)Blecher, Cucurull, Scialom, and Stojnic]{blecher2023nougat}
Lukas Blecher, Guillem Cucurull, Thomas Scialom, and Robert Stojnic.
\newblock Nougat: Neural optical understanding for academic documents.
\newblock \emph{arXiv preprint arXiv:2308.13418}, 2023.

\bibitem[Chen et~al.(2023)Chen, Zhang, Zeng, Zhang, Zhu, and Zhao]{chen2023shikra}
Keqin Chen, Zhao Zhang, Weili Zeng, Richong Zhang, Feng Zhu, and Rui Zhao.
\newblock Shikra: Unleashing multimodal llm's referential dialogue magic.
\newblock \emph{arXiv preprint arXiv:2306.15195}, 2023.

\bibitem[Han et~al.(2023)Han, Zhang, Chen, Yang, Wang, Yu, Fu, and Zhang]{han2023chartllama}
Yucheng Han, Chi Zhang, Xin Chen, Xu Yang, Zhibin Wang, Gang Yu, Bin Fu, and Hanwang Zhang.
\newblock Chartllama: A multimodal llm for chart understanding and generation.
\newblock \emph{arXiv preprint arXiv:2311.16483}, 2023.

\bibitem[Herdade et~al.(2019)Herdade, Kappeler, Boakye, and Soares]{herdade2019image}
Simao Herdade, Armin Kappeler, Kofi Boakye, and Joao Soares.
\newblock Image captioning: Transforming objects into words.
\newblock \emph{Advances in neural information processing systems}, 32, 2019.

\bibitem[Hoque et~al.(2017)Hoque, Setlur, Tory, and Dykeman]{hoque2017applying}
Enamul Hoque, Vidya Setlur, Melanie Tory, and Isaac Dykeman.
\newblock Applying pragmatics principles for interaction with visual analytics.
\newblock \emph{IEEE transactions on visualization and computer graphics}, 24\penalty0 (1):\penalty0 309--318, 2017.

\bibitem[Hoque et~al.(2022)Hoque, Kavehzadeh, and Masry]{hoque2022chart}
Enamul Hoque, Parsa Kavehzadeh, and Ahmed Masry.
\newblock Chart question answering: State of the art and future directions.
\newblock In \emph{Computer Graphics Forum}, pages 555--572. Wiley Online Library, 2022.

\bibitem[Horn(1998)]{horn1998visual}
Robert~E Horn.
\newblock Visual language.
\newblock \emph{MacroVu Inc. Washington}, 1998.

\bibitem[Hsu et~al.(2021)Hsu, Giles, and Huang]{hsu2021scicap}
Ting-Yao Hsu, C~Lee Giles, and Ting-Hao'Kenneth' Huang.
\newblock Scicap: Generating captions for scientific figures.
\newblock \emph{arXiv preprint arXiv:2110.11624}, 2021.

\bibitem[Johnson et~al.(2017)Johnson, Hariharan, Van Der~Maaten, Fei-Fei, Lawrence~Zitnick, and Girshick]{johnson2017clevr}
Justin Johnson, Bharath Hariharan, Laurens Van Der~Maaten, Li Fei-Fei, C Lawrence~Zitnick, and Ross Girshick.
\newblock Clevr: A diagnostic dataset for compositional language and elementary visual reasoning.
\newblock In \emph{Proceedings of the IEEE conference on computer vision and pattern recognition}, pages 2901--2910, 2017.

\bibitem[Kantharaj et~al.(2022{\natexlab{a}})Kantharaj, Do, Leong, Tan, Hoque, and Joty]{kantharaj2022opencqa}
Shankar Kantharaj, Xuan~Long Do, Rixie Tiffany~Ko Leong, Jia~Qing Tan, Enamul Hoque, and Shafiq Joty.
\newblock Opencqa: Open-ended question answering with charts.
\newblock \emph{arXiv preprint arXiv:2210.06628}, 2022{\natexlab{a}}.

\bibitem[Kantharaj et~al.(2022{\natexlab{b}})Kantharaj, Leong, Lin, Masry, Thakkar, Hoque, and Joty]{kantharaj2022chart}
Shankar Kantharaj, Rixie Tiffany~Ko Leong, Xiang Lin, Ahmed Masry, Megh Thakkar, Enamul Hoque, and Shafiq Joty.
\newblock Chart-to-text: A large-scale benchmark for chart summarization.
\newblock \emph{arXiv preprint arXiv:2203.06486}, 2022{\natexlab{b}}.

\bibitem[Kim et~al.(2021)Kim, Hong, Yim, Park, Yim, Hwang, Yun, Han, and Park]{kim2021donut}
Geewook Kim, Teakgyu Hong, Moonbin Yim, Jinyoung Park, Jinyeong Yim, Wonseok Hwang, Sangdoo Yun, Dongyoon Han, and Seunghyun Park.
\newblock Donut: Document understanding transformer without ocr.
\newblock \emph{arXiv preprint arXiv:2111.15664}, 7:\penalty0 15, 2021.

\bibitem[Kim et~al.(2022)Kim, Hong, Yim, Nam, Park, Yim, Hwang, Yun, Han, and Park]{kim2022ocr}
Geewook Kim, Teakgyu Hong, Moonbin Yim, JeongYeon Nam, Jinyoung Park, Jinyeong Yim, Wonseok Hwang, Sangdoo Yun, Dongyoon Han, and Seunghyun Park.
\newblock Ocr-free document understanding transformer.
\newblock In \emph{European Conference on Computer Vision}, pages 498--517. Springer, 2022.

\bibitem[Kingma and Ba(2014)]{kingma2014adam}
Diederik~P Kingma and Jimmy Ba.
\newblock Adam: A method for stochastic optimization.
\newblock \emph{arXiv preprint arXiv:1412.6980}, 2014.

\bibitem[Ko et~al.(2023)Ko, Jeon, Park, Kim, Kim, Kim, and Seo]{ko2023natural}
Hyung-Kwon Ko, Hyeon Jeon, Gwanmo Park, Dae~Hyun Kim, Nam~Wook Kim, Juho Kim, and Jinwook Seo.
\newblock Natural language dataset generation framework for visualizations powered by large language models.
\newblock \emph{arXiv preprint arXiv:2309.10245}, 2023.

\bibitem[Lee et~al.(2023)Lee, Joshi, Turc, Hu, Liu, Eisenschlos, Khandelwal, Shaw, Chang, and Toutanova]{lee2023pix2struct}
Kenton Lee, Mandar Joshi, Iulia~Raluca Turc, Hexiang Hu, Fangyu Liu, Julian~Martin Eisenschlos, Urvashi Khandelwal, Peter Shaw, Ming-Wei Chang, and Kristina Toutanova.
\newblock Pix2struct: Screenshot parsing as pretraining for visual language understanding.
\newblock In \emph{International Conference on Machine Learning}, pages 18893--18912. PMLR, 2023.

\bibitem[Lewis et~al.(2019)Lewis, Liu, Goyal, Ghazvininejad, Mohamed, Levy, Stoyanov, and Zettlemoyer]{lewis2019bart}
Mike Lewis, Yinhan Liu, Naman Goyal, Marjan Ghazvininejad, Abdelrahman Mohamed, Omer Levy, Ves Stoyanov, and Luke Zettlemoyer.
\newblock Bart: Denoising sequence-to-sequence pre-training for natural language generation, translation, and comprehension.
\newblock \emph{arXiv preprint arXiv:1910.13461}, 2019.

\bibitem[Li et~al.(2023)Li, Li, Savarese, and Hoi]{li2023blip}
Junnan Li, Dongxu Li, Silvio Savarese, and Steven Hoi.
\newblock Blip-2: Bootstrapping language-image pre-training with frozen image encoders and large language models.
\newblock \emph{arXiv preprint arXiv:2301.12597}, 2023.

\bibitem[Li and Tajbakhsh(2023)]{li2023scigraphqa}
Shengzhi Li and Nima Tajbakhsh.
\newblock Scigraphqa: A large-scale synthetic multi-turn question-answering dataset for scientific graphs.
\newblock \emph{arXiv preprint arXiv:2308.03349}, 2023.

\bibitem[Lin et~al.(2023)Lin, Liu, Zhang, Gao, Qiu, Xiao, Qiu, Lin, Shao, Chen, et~al.]{lin2023sphinx}
Ziyi Lin, Chris Liu, Renrui Zhang, Peng Gao, Longtian Qiu, Han Xiao, Han Qiu, Chen Lin, Wenqi Shao, Keqin Chen, et~al.
\newblock Sphinx: The joint mixing of weights, tasks, and visual embeddings for multi-modal large language models.
\newblock \emph{arXiv preprint arXiv:2311.07575}, 2023.

\bibitem[Liu et~al.(2022{\natexlab{a}})Liu, Eisenschlos, Piccinno, Krichene, Pang, Lee, Joshi, Chen, Collier, and Altun]{liu2022deplot}
Fangyu Liu, Julian~Martin Eisenschlos, Francesco Piccinno, Syrine Krichene, Chenxi Pang, Kenton Lee, Mandar Joshi, Wenhu Chen, Nigel Collier, and Yasemin Altun.
\newblock Deplot: One-shot visual language reasoning by plot-to-table translation.
\newblock \emph{arXiv preprint arXiv:2212.10505}, 2022{\natexlab{a}}.

\bibitem[Liu et~al.(2022{\natexlab{b}})Liu, Piccinno, Krichene, Pang, Lee, Joshi, Altun, Collier, and Eisenschlos]{liu2022matcha}
Fangyu Liu, Francesco Piccinno, Syrine Krichene, Chenxi Pang, Kenton Lee, Mandar Joshi, Yasemin Altun, Nigel Collier, and Julian~Martin Eisenschlos.
\newblock Matcha: Enhancing visual language pretraining with math reasoning and chart derendering.
\newblock \emph{arXiv preprint arXiv:2212.09662}, 2022{\natexlab{b}}.

\bibitem[Liu et~al.(2023{\natexlab{a}})Liu, Wang, Yao, Chen, Song, Cho, Yacoob, and Yu]{liu2023mmc}
Fuxiao Liu, Xiaoyang Wang, Wenlin Yao, Jianshu Chen, Kaiqiang Song, Sangwoo Cho, Yaser Yacoob, and Dong Yu.
\newblock Mmc: Advancing multimodal chart understanding with large-scale instruction tuning.
\newblock \emph{arXiv preprint arXiv:2311.10774}, 2023{\natexlab{a}}.

\bibitem[Liu et~al.(2023{\natexlab{b}})Liu, Li, Wu, and Lee]{liu2023visual}
Haotian Liu, Chunyuan Li, Qingyang Wu, and Yong~Jae Lee.
\newblock Visual instruction tuning.
\newblock \emph{arXiv preprint arXiv:2304.08485}, 2023{\natexlab{b}}.

\bibitem[Liu et~al.(2021)Liu, Lin, Cao, Hu, Wei, Zhang, Lin, and Guo]{liu2021swin}
Ze Liu, Yutong Lin, Yue Cao, Han Hu, Yixuan Wei, Zheng Zhang, Stephen Lin, and Baining Guo.
\newblock Swin transformer: Hierarchical vision transformer using shifted windows.
\newblock In \emph{Proceedings of the IEEE/CVF international conference on computer vision}, pages 10012--10022, 2021.

\bibitem[Lv et~al.(2023)Lv, Huang, Chen, Cui, Ma, Chang, Huang, Wang, Dong, Luo, et~al.]{lv2023kosmos}
Tengchao Lv, Yupan Huang, Jingye Chen, Lei Cui, Shuming Ma, Yaoyao Chang, Shaohan Huang, Wenhui Wang, Li Dong, Weiyao Luo, et~al.
\newblock Kosmos-2.5: A multimodal literate model.
\newblock \emph{arXiv preprint arXiv:2309.11419}, 2023.

\bibitem[Masry et~al.(2022)Masry, Long, Tan, Joty, and Hoque]{masry2022chartqa}
Ahmed Masry, Do~Xuan Long, Jia~Qing Tan, Shafiq Joty, and Enamul Hoque.
\newblock Chartqa: A benchmark for question answering about charts with visual and logical reasoning.
\newblock \emph{arXiv preprint arXiv:2203.10244}, 2022.

\bibitem[Masry et~al.(2023)Masry, Kavehzadeh, Do, Hoque, and Joty]{masry2023unichart}
Ahmed Masry, Parsa Kavehzadeh, Xuan~Long Do, Enamul Hoque, and Shafiq Joty.
\newblock Unichart: A universal vision-language pretrained model for chart comprehension and reasoning.
\newblock \emph{arXiv preprint arXiv:2305.14761}, 2023.

\bibitem[Methani et~al.(2020)Methani, Ganguly, Khapra, and Kumar]{methani2020plotqa}
Nitesh Methani, Pritha Ganguly, Mitesh~M Khapra, and Pratyush Kumar.
\newblock Plotqa: Reasoning over scientific plots.
\newblock In \emph{Proceedings of the IEEE/CVF Winter Conference on Applications of Computer Vision}, pages 1527--1536, 2020.

\bibitem[Oquab et~al.(2023)Oquab, Darcet, Moutakanni, Vo, Szafraniec, Khalidov, Fernandez, Haziza, Massa, El-Nouby, et~al.]{oquab2023dinov2}
Maxime Oquab, Timoth{\'e}e Darcet, Th{\'e}o Moutakanni, Huy Vo, Marc Szafraniec, Vasil Khalidov, Pierre Fernandez, Daniel Haziza, Francisco Massa, Alaaeldin El-Nouby, et~al.
\newblock Dinov2: Learning robust visual features without supervision.
\newblock \emph{arXiv preprint arXiv:2304.07193}, 2023.

\bibitem[Ouyang et~al.(2022)Ouyang, Wu, Jiang, Almeida, Wainwright, Mishkin, Zhang, Agarwal, Slama, Ray, et~al.]{ouyang2022training}
Long Ouyang, Jeffrey Wu, Xu Jiang, Diogo Almeida, Carroll Wainwright, Pamela Mishkin, Chong Zhang, Sandhini Agarwal, Katarina Slama, Alex Ray, et~al.
\newblock Training language models to follow instructions with human feedback.
\newblock \emph{Advances in Neural Information Processing Systems}, 35:\penalty0 27730--27744, 2022.

\bibitem[Radford et~al.(2021)Radford, Kim, Hallacy, Ramesh, Goh, Agarwal, Sastry, Askell, Mishkin, Clark, et~al.]{radford2021learning}
Alec Radford, Jong~Wook Kim, Chris Hallacy, Aditya Ramesh, Gabriel Goh, Sandhini Agarwal, Girish Sastry, Amanda Askell, Pamela Mishkin, Jack Clark, et~al.
\newblock Learning transferable visual models from natural language supervision.
\newblock In \emph{International conference on machine learning}, pages 8748--8763. PMLR, 2021.

\bibitem[Raffel et~al.(2020)Raffel, Shazeer, Roberts, Lee, Narang, Matena, Zhou, Li, and Liu]{raffel2020exploring}
Colin Raffel, Noam Shazeer, Adam Roberts, Katherine Lee, Sharan Narang, Michael Matena, Yanqi Zhou, Wei Li, and Peter~J Liu.
\newblock Exploring the limits of transfer learning with a unified text-to-text transformer.
\newblock \emph{The Journal of Machine Learning Research}, 21\penalty0 (1):\penalty0 5485--5551, 2020.

\bibitem[Rahman et~al.(2022)Rahman, Hasan, and Farhad]{rahman2022chartsumm}
Raian Rahman, Rizvi Hasan, and Abdullah~Al Farhad.
\newblock \emph{ChartSumm: A large scale benchmark for Chart to Text Summarization}.
\newblock PhD thesis, Department of Computer Science and Engineering (CSE), Islamic University of~…, 2022.

\bibitem[Shao et~al.(2023)Shao, Hu, Gao, Lei, Zhang, Meng, Xu, Huang, Li, Qiao, et~al.]{shao2023tiny}
Wenqi Shao, Yutao Hu, Peng Gao, Meng Lei, Kaipeng Zhang, Fanqing Meng, Peng Xu, Siyuan Huang, Hongsheng Li, Yu Qiao, et~al.
\newblock Tiny lvlm-ehub: Early multimodal experiments with bard.
\newblock \emph{arXiv preprint arXiv:2308.03729}, 2023.

\bibitem[Tang et~al.(2023)Tang, Boggust, and Satyanarayan]{tang2023vistext}
Benny~J Tang, Angie Boggust, and Arvind Satyanarayan.
\newblock Vistext: A benchmark for semantically rich chart captioning.
\newblock \emph{arXiv preprint arXiv:2307.05356}, 2023.

\bibitem[Touvron et~al.(2023)Touvron, Lavril, Izacard, Martinet, Lachaux, Lacroix, Rozi{\`e}re, Goyal, Hambro, Azhar, et~al.]{touvron2023llama}
Hugo Touvron, Thibaut Lavril, Gautier Izacard, Xavier Martinet, Marie-Anne Lachaux, Timoth{\'e}e Lacroix, Baptiste Rozi{\`e}re, Naman Goyal, Eric Hambro, Faisal Azhar, et~al.
\newblock Llama: Open and efficient foundation language models.
\newblock \emph{arXiv preprint arXiv:2302.13971}, 2023.

\bibitem[Vinyals et~al.(2015)Vinyals, Toshev, Bengio, and Erhan]{vinyals2015show}
Oriol Vinyals, Alexander Toshev, Samy Bengio, and Dumitru Erhan.
\newblock Show and tell: A neural image caption generator.
\newblock In \emph{Proceedings of the IEEE conference on computer vision and pattern recognition}, pages 3156--3164, 2015.

\bibitem[Wei et~al.(2022)Wei, Wang, Schuurmans, Bosma, Xia, Chi, Le, Zhou, et~al.]{wei2022chain}
Jason Wei, Xuezhi Wang, Dale Schuurmans, Maarten Bosma, Fei Xia, Ed Chi, Quoc~V Le, Denny Zhou, et~al.
\newblock Chain-of-thought prompting elicits reasoning in large language models.
\newblock \emph{Advances in Neural Information Processing Systems}, 35:\penalty0 24824--24837, 2022.

\bibitem[Woo et~al.(2023)Woo, Debnath, Hu, Chen, Liu, Kweon, and Xie]{woo2023convnext}
Sanghyun Woo, Shoubhik Debnath, Ronghang Hu, Xinlei Chen, Zhuang Liu, In~So Kweon, and Saining Xie.
\newblock Convnext v2: Co-designing and scaling convnets with masked autoencoders.
\newblock In \emph{Proceedings of the IEEE/CVF Conference on Computer Vision and Pattern Recognition}, pages 16133--16142, 2023.

\bibitem[Xia et~al.(2023)Xia, Zhang, Peng, Liao, Ye, Shi, Yan, and Qiao]{xia2023structchart}
Renqiu Xia, Bo Zhang, Haoyang Peng, Ning Liao, Peng Ye, Botian Shi, Junchi Yan, and Yu Qiao.
\newblock Structchart: Perception, structuring, reasoning for visual chart understanding.
\newblock \emph{arXiv preprint arXiv:2309.11268}, 2023.

\bibitem[Xu et~al.(2023)Xu, Shao, Zhang, Gao, Liu, Lei, Meng, Huang, Qiao, and Luo]{xu2023lvlm}
Peng Xu, Wenqi Shao, Kaipeng Zhang, Peng Gao, Shuo Liu, Meng Lei, Fanqing Meng, Siyuan Huang, Yu Qiao, and Ping Luo.
\newblock Lvlm-ehub: A comprehensive evaluation benchmark for large vision-language models.
\newblock \emph{arXiv preprint arXiv:2306.09265}, 2023.

\bibitem[Yang et~al.(2023)Yang, Zhang, Li, Zou, Li, and Gao]{yang2023set}
Jianwei Yang, Hao Zhang, Feng Li, Xueyan Zou, Chunyuan Li, and Jianfeng Gao.
\newblock Set-of-mark prompting unleashes extraordinary visual grounding in gpt-4v.
\newblock \emph{arXiv preprint arXiv:2310.11441}, 2023.

\bibitem[Zhang et~al.(2023{\natexlab{a}})Zhang, Fei, Yao, Ji, Li, Liu, and Chua]{zhang2023transfer}
Ao Zhang, Hao Fei, Yuan Yao, Wei Ji, Li Li, Zhiyuan Liu, and Tat-Seng Chua.
\newblock Transfer visual prompt generator across llms.
\newblock \emph{arXiv preprint arXiv:2305.01278}, 2023{\natexlab{a}}.

\bibitem[Zhang et~al.(2023{\natexlab{b}})Zhang, Han, Zhou, Hu, Yan, Lu, Li, Gao, and Qiao]{zhang2023llama}
Renrui Zhang, Jiaming Han, Aojun Zhou, Xiangfei Hu, Shilin Yan, Pan Lu, Hongsheng Li, Peng Gao, and Yu Qiao.
\newblock Llama-adapter: Efficient fine-tuning of language models with zero-init attention.
\newblock \emph{arXiv preprint arXiv:2303.16199}, 2023{\natexlab{b}}.

\bibitem[Zhang et~al.(2023{\natexlab{c}})Zhang, Sun, Chen, Xiao, Shao, Zhang, Chen, and Luo]{zhang2023gpt4roi}
Shilong Zhang, Peize Sun, Shoufa Chen, Min Xiao, Wenqi Shao, Wenwei Zhang, Kai Chen, and Ping Luo.
\newblock Gpt4roi: Instruction tuning large language model on region-of-interest.
\newblock \emph{arXiv preprint arXiv:2307.03601}, 2023{\natexlab{c}}.

\bibitem[Zhou et~al.(2023)Zhou, Fung, Chen, Thomas, Ji, and Chang]{zhou2023enhanced}
Mingyang Zhou, Yi~R Fung, Long Chen, Christopher Thomas, Heng Ji, and Shih-Fu Chang.
\newblock Enhanced chart understanding in vision and language task via cross-modal pre-training on plot table pairs.
\newblock \emph{arXiv preprint arXiv:2305.18641}, 2023.

\bibitem[Zhu et~al.(2023)Zhu, Chen, Shen, Li, and Elhoseiny]{zhu2023minigpt}
Deyao Zhu, Jun Chen, Xiaoqian Shen, Xiang Li, and Mohamed Elhoseiny.
\newblock Minigpt-4: Enhancing vision-language understanding with advanced large language models.
\newblock \emph{arXiv preprint arXiv:2304.10592}, 2023.

\end{thebibliography}
